%% file: main.tex
\documentclass[10pt,twocolumn,letterpaper]{article}

\usepackage[pagenumbers]{cvpr} 
\usepackage{svg}


\definecolor{cvprblue}{rgb}{0.21,0.49,0.74}
\usepackage[pagebackref,breaklinks,colorlinks,allcolors=cvprblue]{hyperref}
\usepackage{amssymb,amsmath}

\def\paperID{*****} 
\def\confName{CVPR}
\def\confYear{2025}

\title{Improving Object Detection by Modifying Synthetic Data with Explainable AI} 

\author{Nitish Mital$^{1,*}$, 
	Simon Malzard$^{1}$,
	Richard Walters$^{1}$, \,
	Celso M. De Melo$^{2}$, \,
	Raghuveer Rao$^{2}$, \\ and
	Victoria Nockles$^{1}$ \\
	\vspace{-0.1cm}
	$^{1}$DARe, The Alan Turing Institute, London NW1 2DB, UK.\\
	$^{2}$DEVCOM Army Research Laboratory, USA. \\
        $^{*}${\tt\small nmital@turing.ac.uk}
}

\begin{document}
\maketitle
\input{sec_iccv/0_abstract}    
\input{sec_iccv/1_intro}

\input{sec_iccv/2_method}
\input{sec_iccv/3_experiments}
\input{sec_iccv/4_discussion}
{
    \small
    \bibliographystyle{ieeenat_fullname}
    \bibliography{main}
}
\clearpage
\input{sec_iccv/X_suppl}

\end{document}

%% file: sec_iccv/0_abstract.tex
\begin{abstract}

Limited real-world data severely impacts model performance in many computer vision domains, particularly for samples that are underrepresented in training.
Synthetically generated images are a promising solution, but 1) it remains unclear how to design synthetic training data to optimally improve model performance (e.g, whether and where to introduce more realism or more abstraction) and 2) the domain expertise, time and effort required from human operators for this design and optimisation process represents a major practical challenge.
Here we propose a novel conceptual approach to improve the efficiency of designing synthetic images, by using robust Explainable AI (XAI) techniques to guide a human-in-the-loop process of modifying 3D mesh models used to generate these images. Importantly, this framework allows both modifications that increase and decrease realism in synthetic data, which can both improve model performance. 
We illustrate this concept using a real-world example where data are sparse; detection of vehicles in infrared imagery. We fine-tune an initial YOLOv8 model on the ATR DSIAC infrared dataset and synthetic images generated from 3D mesh models in the Unity gaming engine, and then use XAI saliency maps to guide modification of our Unity models. We show that synthetic data can improve detection of vehicles in orientations unseen in training by 4.6\% (to mAP50 = 94.6\%). We further improve performance by an additional 1.5\% (to 96.1\%) through our new XAI-guided approach, which reduces misclassifications through both increasing and decreasing the realism of different parts of the synthetic data.  
Our proof-of-concept results pave the way for fine, XAI-controlled curation of synthetic datasets tailored to improve object detection performance, whilst simultaneously reducing the burden on human operators in designing and optimising these datasets.\end{abstract}

%% file: sec_iccv/1_intro.tex
\section{Introduction}


In many machine learning domains the collection of sufficient real-world data is challenging and can severely impact model performance, particularly when running inference on test samples and scenarios that are unseen or strongly under-represented in training data. Barriers to data collection include time, cost, safety, complexity, and legal or privacy restrictions \cite{singh2021rise}, and the defence \cite{huang2023synthetic}, healthcare \cite{giuffre2023harnessing}, finance \cite{assefa2020generating}, and autonomous systems \cite{deng2023exploring} domains frequently face strict regulations that limit sharing of sensitive data. For computer vision models, synthetic training images provide a promising solution \cite{de2022next}, and such data can be generated by a range of techniques including synthetic composite imagery \cite{sakaridis2018semantic,varol2017learning,shermeyer2021rareplanes}, GANs \cite{zhu2017unpaired,choi2018stargan,isola2017image,karras2019style}, Diffusion models \cite{ramesh2021zero,rombach2022high}, GAN-Diffusion hybrids \cite{wang2022diffusion} and 3D rendered environments \cite{qiu2016unrealcv,qiu2017unrealcv,borkman2021unity}. 

However, despite significant progress in the generation of synthetic data, it remains unclear how best to design synthetic data to optimally improve model performance, i.e. what constitutes useful or optimal synthetic data to augment a real-world dataset for a given task. For example, synthetic data can be subject to a range of issues such as the domain gap between real and synthetic samples \cite{9266246,bai2024bridging}, poor generalisability owing to lack of diversity \cite{PASANISI2023816,tremblay2018training,grushko2023hadr}, and mode collapse \cite{shumailov2023curse,9934291,kodali2017convergence}, but approaches to mitigate these issues range from increasing realism in synthetic data \cite{movshovitz2016useful,kornfein2023closing}, to making data less realistic in order to disrupt sources of misclassification and force a network to learn the essential features of an object (as in domain randomisation \cite{tremblay2018training}). Therefore even at a fundamental level, it is unclear when more realism or when less realism is beneficial in synthetic data. Furthermore, whilst human-in-the-loop  design and optimisation of synthetic datasets is desirable as it embeds human knowledge into the training process, improving model performance while also enhancing its robustness and generalisation \cite{WU2022364}, it requires significant domain expertise, time and effort from human operators. As these operators are already burdened with intense evaluation tasks (identifying failure modes, resolving them by annotating instances, curating the dataset, changing the training algorithm) this additional workload represents a major practical barrier to the optimal use of synthetic data \cite{10530996}.

%




One approach to simultaneously address both these issues is to let model performance directly tell the human operator where and what type of modifications are needed in the synthetic data, which could include increases or decreases in data realism as required.
To this aim, we propose a novel explainable synthetic data generation framework with human-in-the-loop for object detection, using insights from Explainable AI (XAI) techniques to efficiently guide this process. Robust XAI techniques (SHAP \cite{lundbergunified}) are used to identify features that cause confusion or distinguish between classes, and the human operator then modifies the synthetic training dataset to respectively disrupt or reinforce these features, so as to steer the model's attention and improve performance.  



\noindent{}\textbf{Related Work.} Previous studies have developed techniques to test and `debug' models by identifying low performing subsets of data using human-in-the-loop interactions \cite{10378324, Ganguli2022RedTL, ribeiro-lundberg-2022-adaptive, checklist:acl20, wiles2022discovering}, and then by fine-tuning models to improve their performance on these ``hard samples''.  In our approach, we instead use SHAP XAI \cite{lundbergunified} to identify features of the hard samples causing low model performance, and then carefully modify these features in the synthetic training data in order to improve performance on these hard samples.

Whilst previous works have have either improved computer vision performance using large volumes of synthetic data \cite{sabirsynthetic,maria2023ets2,shen2021automatic}, or have investigated computer vision explainability \cite{10022096,kuroki2024bsed}, our use of XAI to guide improvements in generation of new synthetic data is novel. Saliency-guided data augmentation has been used to mix and modify training samples \cite{Regulariza2020SaliencyMixAS}, but we take a fundamentally different approach by directly modifying mesh models within a gaming engine (Unity3D \cite{unity}) that provides fine-grained control of textures, lighting and materials that, to the best of our knowledge, is not achievable with either this saliency-guided data augmentation \cite{Regulariza2020SaliencyMixAS} or with current controlled diffusion models \cite{Gu2024KaleidoDI, 10203593, 10.1007}.


A close analogue to our approach is ``feature steering'' which identifies specific features using mechanistic interpretability \cite{bereska2024mechanistic}, and weakens or strengthens their activation to influence model outputs. However this approach \cite{templeton2024scaling} does not change the underlying model weights and requires access to the complete model, whilst our method is model-agnostic and instead improves performance by steering model weights using targeted modifications to synthetic data. Our method also differs to other methods such as using a varifocal loss \cite{9578034} or online hard-example mining \cite{Shrivastava2016TrainingRO}, by directly addressing the limitations of the dataset itself rather than optimising the training procedure alone.

In this paper we use a real-world example of detection of vehicles in infrared data to demonstrate our new approach; for this task collection of data is challenging and synthetic data are of especial benefit for detection of unseen scenarios. We use the Defense Systems Information Analysis Centre Automated Target Recognition (DSIAC ATR) dataset to show how our XAI-guided approach to synthetic data modification can improve detection of vehicles in orientations unseen in real-world training data. 
Our contributions are:

\begin{itemize}
\item{We propose a novel human-in-the-loop framework using explainable-AI techniques to guide modification of synthetic data, in order to improve performance of an object detection model trained with these data and reduce workload for the human operator. }
\item{We show that XAI-guided modification of synthetic data, which involves both increasing and decreasing data realism, improves performance by up to 1.5\% over an initial 4.6\% gain from using a base version of synthetic data.}
\end{itemize}

%% file: sec_iccv/2_method.tex
\section{Method}
\label{sec:method}
\begin{figure*}[ht]
\centering
\includegraphics[width=0.8\textwidth]{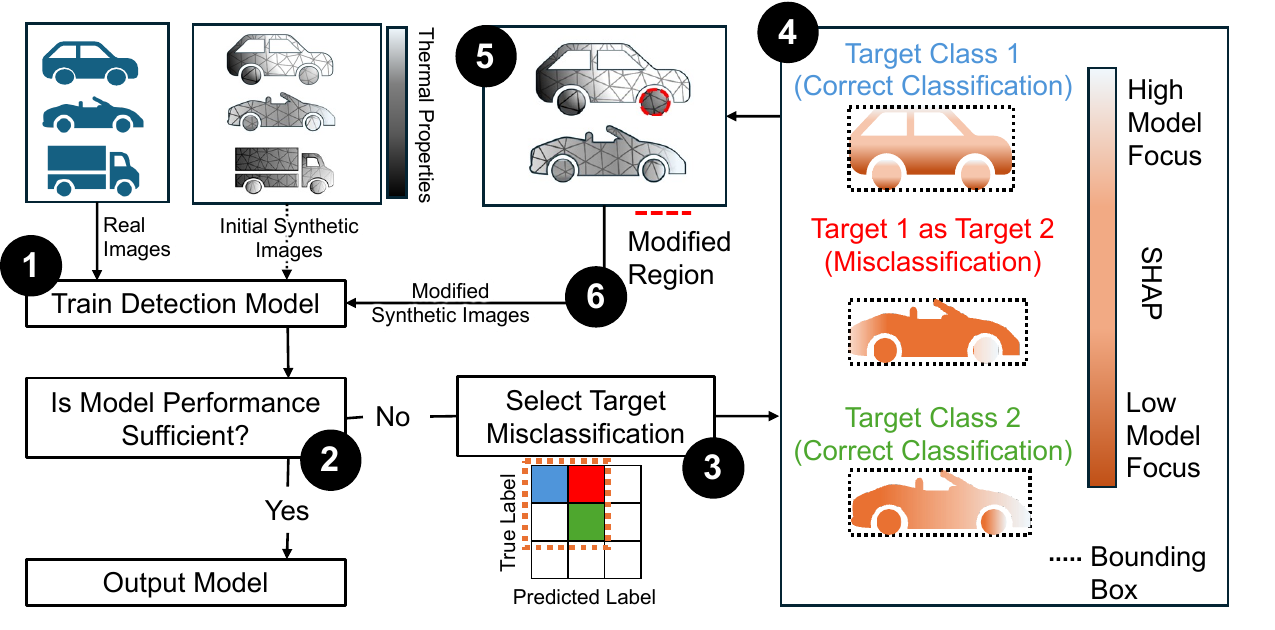}
\caption{Conceptual illustration of our proposed approach for improving the performance of object detection and classification algorithms that are trained on synthetic images, through use of SHAP saliency maps to guide modification of mesh models used to generate synthetic data. Numbered circles represent the six key steps in our approach outlined in Section \ref{sec:procedure} and demonstrated using examples in Section \ref{sec:XAI_examples}. 
}
\label{fig:toymodel}
\end{figure*}


We propose a method to improve the performance of object detection and classification algorithms that are trained on synthetic images, through using saliency maps to guide modification of these synthetic data. 
We first motivate our approach with a mathematical toy model that illustrates how characteristics of training data can affect model solutions.
In particular, we introduce and explain the \textit{common features} and \textit{unique features} for a pair of classes, which are key to understanding the theoretical rationale for our methodology. 

For our toy model, we consider an MLP binary classifier for classes $k$ and $k'$, that classifies an input sample $\mathbf{x}_{j}$, consisting of $N$ features, $\mathbf{v}_{1}, \ldots, \mathbf{v}_{N}$, from class $j\in \{k, k'\}$, and assigns a confidence score for class $k$ via a classification layer as $f_k(\mathbf{x}_j) \triangleq \sum_{i=1}^N w^k_i \phi(\mathbf{x}_j,\mathbf{v}_i) \triangleq \boldsymbol{\phi}(\mathbf{x}_j)^T \mathbf{w}_k$. Here, $\phi(\mathbf{x}_j, \mathbf{v}_i)$ is a quantification of the extent of the presence of feature $\mathbf{v}_i$ in $\mathbf{x}_j$. By training the classification layer $\mathbf{w}_k$ to simultaneously maximise the expected confidence of all correct classifications of class $k$, whilst also minimising the expected confidence for all misclassifications of $\mathbf{x}_{k'}$ to class $k$, we get optimal weights across the whole dataset given by: $\mathbf{w}_k^* \propto \mathbb{E} [\boldsymbol{\phi}(\mathbf{x}_k)] - \mathbb{E}[\boldsymbol{\phi} (\mathbf{x}_{k'})].$ 

Therefore, in our trained binary classifier, large weights (large values in $\mathbf{w}_k^*$) are assigned to \textit{unique features} which are prominent in class $k$; i.e., features $\mathbf{v}_i$ for which $\mathbb{E}[\phi(\mathbf{x}_k, \mathbf{v}_i)]$ is high, but are less prominent in the other class $k'$. 
Conversely, small weights (small values in $\mathbf{w}_k^*$) are assigned to \textit{common features}, those that are prominent in both classes. 
As such, misclassifications between classes are more likely to arise when unique features have weights that are too small, and when common features have weights that are too large. Although this toy problem is simple, we suggest that this concept is more generally applicable to deep learning classification models. Our proposed procedure focuses on identifying unique and common features and then reinforcing and disrupting them respectively. In order to reduce confusion between two classes, we make modifications to the mesh models of these classes to promote unique features in each model (which we term \textit{Reinforcing modification}) or to suppress common features associated with misclassification (which we term \textit{Disruptive modification}). 

\noindent\textit{Reinforcing modification:} In the framework of our toy MLP classification model, promoting unique features is represented by increasing $\mathbb{E}[\phi(\mathbf{x}_k, \mathbf{v}_i)]$ for a unique feature $\mathbf{v}_i$, and can be achieved by introducing synthetic samples of class $k$ in which feature $\mathbf{v}_i$ is prominent. This modification steers the model to assign a higher weight to this feature. We call this a \textit{Reinforcing} modification. In this way, an outlier/edge sample from class $k$ that has low prominence of unique features will be less likely to be misclassified.
In practice, this Reinforcing modification in a mesh model should involve making these features in the synthetic data more similar to those in the real samples, or more prominent than those causing confusion (e.g. by comparative exaggeration).
    
\noindent\textit{Disruptive modification:} Alternatively, decreasing $\mathbb{E}[\phi(\mathbf{x}_k, \mathbf{v}_j)]$ for a common feature $\mathbf{v}_j$, by introducing synthetic samples of class $k$ in which feature $\mathbf{v}_j$ is absent or less prominent, steers the model to assign a lower weight to $\mathbf{v}_j$. We call this a \textit{Disruptive} modification, and an outlier/edge sample from class $k$ that has high prominence of common features will be less likely to be misclassified.
In practice, this Disruptive modification in a mesh model should involve reducing this feature, which could involve increasing it's diversity/variety or making synthetic data less realistic in the region of this model (e.g. as in domain randomisation).

It is important to note that although here we predominantly focus on reinforcing through increased realism and disruption through decreased realism, reinforcing modification could equally involve decreased realism (e.g. exaggeration of unique features) and disruptive modification could involve increased realism (e.g. greater, more realistic variety of common features).

\subsection{Procedure}
\label{sec:procedure}

Our proposed human-in-the-loop procedure, which exploits this ability to reinforce unique features and disrupt common features to reduce misclassifications and improve model performance, is laid out in the following six steps, which are also illustrated in Fig. \ref{fig:toymodel} for a simple graphical example.

    \noindent{}\textbf{(1) Human operator trains an object detection model} on initial real and synthetic images.
    
    \noindent{}\textbf{(2) Operator evaluates the performance of the model}, and terminates procedure if it is sufficient for requirements.
    
    \noindent{}\textbf{(3) Operator identifies a target misclassification from the model confusion matrix}, typically the largest off-diagonal, non-background element (red square in Fig. \ref{fig:toymodel}), and associated classes (blue=hatchback, green=sports car).
    
    \noindent{}\textbf{(4) Operator identifies cause of the misclassification}, by calculating SHAP saliency contributions (more detail below) for the correct classifications of both classes and for the misclassification. Operator identifies unique features that distinguish between the two classes and common features that cause confusion, by examining correlations in the location, pattern and size of the SHAP values between the correct classifications and the misclassification.
    
    \noindent{}\textbf{(5) Operator makes modifications to the mesh models, guided by SHAP comparisons,} to reinforce unique features or disrupt common features to respectively increase or decrease their prominence.
    
    \noindent{}\textbf{(6) Modified mesh models are used to generate new version of the synthetic dataset, and process reverts to step 1., where operator retrains the model.} The process is repeated until sufficient model performance criteria are met. 

It is important to note that this approach is general and could equally apply to any detection task where synthetic imagery can be generated from 3D mesh models. 

We now provide additional details on our use of XAI. We use SHAP \cite{lundbergunified} because it is model-agnostic and satisfies desirable theoretical properties for explainability \cite{molnar2022} over alternatives such as attention \cite{Jain2019AttentionIN}.
SHAP values \cite{lundbergunified}, which are an approximation to Shapley values \cite{shapley1953value}, take an input set of superpixels within the bounding box for each image and attribute the marginal contributions of each superpixel to a specific classification. To calculate the superpixel attributions for each sample, a random mask is applied over the image, where masked portions of the sample are replaced with samples from a background set. Inference is then computed over this masked image and each pixel assigned its classification score. This is repeated for the same image over $1000$ different random masks to find the average contribution of a given pixel towards a specific classification, using the SHAP Kernel method \cite{lundbergunified}. 
We compute pixel attributions over $50$ randomly selected test samples for which the model predicts high confidence for the specified class. For each pixel, we count the number of samples in which it has a contribution greater than a user-defined threshold (we use $40\%$ of the highest contribution in each sample, balancing a trade-off between information and noise). Thus we obtain an average SHAP contribution map within the bounding box, where each pixel is assigned a score of the number of samples in which it contributed more than the given threshold. In order to better visualise the average SHAP contribution maps, we mask out pixels that have a score less than another user-defined threshold (we use $50\%$ of the highest score), and superimpose this on a representative test image to identify parts of the target that contribute most to the specific classification (see purple masks on Model Focus images in Fig. \ref{fig:shap}).


Next we detail how we use SHAP contribution maps in practice to identify what area of the mesh model to modify, and to determine what type of change is needed (steps 4 and 5 in the previous section).
We plot the average SHAP contribution maps inside the bounding boxes for samples in the two target classes (for the case where class A is misclassifed as class B). Three saliency maps are plotted; for the correct classification of class A, for the correct classification of class B, and the saliency map corresponding to the misclassification of class A as class B. Both unique and common features can be identified by comparison of these three SHAP saliency maps, with reference to the ground truth models. If the SHAP maps for the misclassification (class A as class B) and correct classification of class B both have high saliency in the same location \textit{and} the two datasets used to calculate these SHAP images are similar (i.e. have similar brightness, patterns) in this same region, then this is a common feature. Conversely, if a high saliency pattern at a location in the correct classification bounding box does not appear with any significant saliency in the misclassification at the same location, then this is a unique feature. It is important to note that this comparison must also account for variation across each class. For this reason we first need to cluster validation samples, using dimensional reduction techniques (e.g. principal component analysis), or using appropriate data labels corresponding to a natural dimension of variation (e.g. vehicle orientation in our example). Comparison of SHAP images is then carried out across each cluster.

Reinforcing unique features can be achieved by introducing synthetic samples of class A that make these features and patterns in the synthetic data more similar to those in the real samples, which in future work could be measured by perceptual image quality metrics (e.g. SSIM \cite{1284395}, MSSIM \cite{1292216}, LPIPS \cite{8578166}). Disrupting common features involves introducing synthetic samples where these features or patterns are absent or faint, therefore making the corresponding regions in the synthetic and real samples dissimilar.

\section{Dataset}
\label{sec:Dataset}

\begin{figure}[ht]
    \centering
    \includegraphics[width=0.45\textwidth]{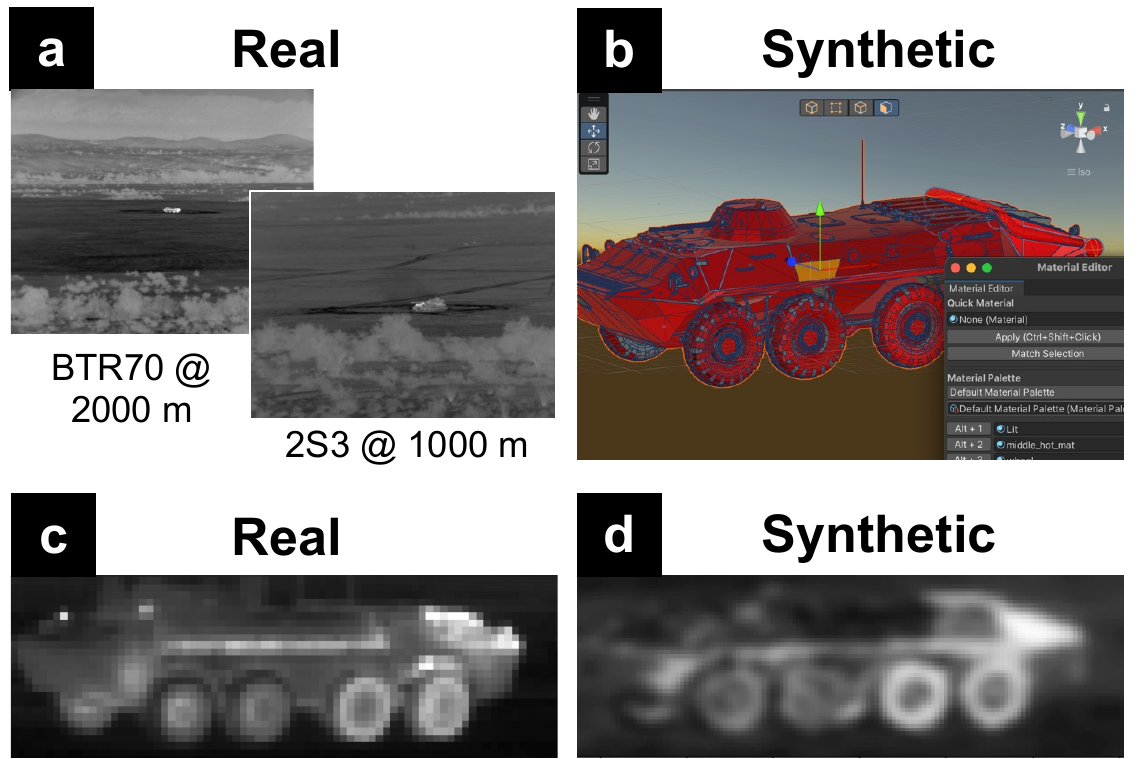}
    \caption{Illustration of the DSIAC ATR Dataset and Synthetic images. (a) Real example MWIR images showing different vehicle classes (BTR70 and 2S3) at different ranges ($2000m$ and $1000m$). (b) Screenshot of the Probuilder tool \cite{probuilder} in Unity for the BTR70 class, where we set material properties of all the faces of the object mesh in order to generate synthetic MWIR images. Comparison of real and synthetic MWIR images for the BTR70 are shown in panels (c) and (d), respectively.}
\label{Fig:Dataset}
\end{figure}

\noindent{}\textbf{The ATR DSIAC dataset} \cite{DSIAC} was collected by the former US Army Night Vision and Electronic Sensors Directorate (NVESD). The dataset contains mid-wave infrared (MWIR, shortened to IR from here on) and electro-optical videos of $8$ different vehicle classes (Pickup Truck, SUV, BTR70, BRDM2, BMP2, T72, ZSU23-4, and 2S3) taken at daytime and night time at distances of $1000$m, $1500$m, and $2000$m (see Fig. \ref{Fig:Dataset}a and Fig. 1, Section 1 in supplementary material for examples of complete frames, closeups of all classes, and more details). In the videos the vehicles move in approximately circular paths, and the camera is located at a slightly elevated side-on position, so that a range of orientations (front, back, sides) are captured in the dataset. We extract video frames as image samples and only use night-time images to avoid rendering complexities caused by solar illumination and material reflectance when generating our synthetic dataset. 

\noindent{}\textbf{Synthetic data generation.} To generate synthetic data, we use the Unity3D gaming engine along with 3D models of the vehicle classes that are freely available online. We use these to develop a 3D scene with similar distance ranges and camera resolutions to the real IR images in the ATR DSIAC dataset. To simulate infrared image capture, we use the Probuilder tool \cite{probuilder} in Unity to set material properties (reflectance and emission) of the face of each mesh element, as illustrated by the selected orange panel in Fig. \ref{fig:train-test-data}b. 
We set the material properties and the scene details to produce renderings with roughly similar pixel values of the vehicles as in the original images from the dataset, and we implement a 2D convolution of the rendered image with an Airy-disk kernel to simulate the diffraction effect from a circular pixel aperture. 
In future it would be possible to automate this procedure, including setting of material properties and other parameters, but for this proof of concept we do this manually; a key benefit of our method is that it does not require photorealistic synthetic data due to the XAI-guided procedure, which focuses on performance gain from synthetic data alone, and therefore may modify synthetic data to be less realistic as well as more realistic.

%% file: sec_iccv/3_experiments.tex
\section{Experimental Setup}
\label{sec:experiments}

\begin{figure*}[ht]
    \centering
    \includegraphics[width=0.95\textwidth]{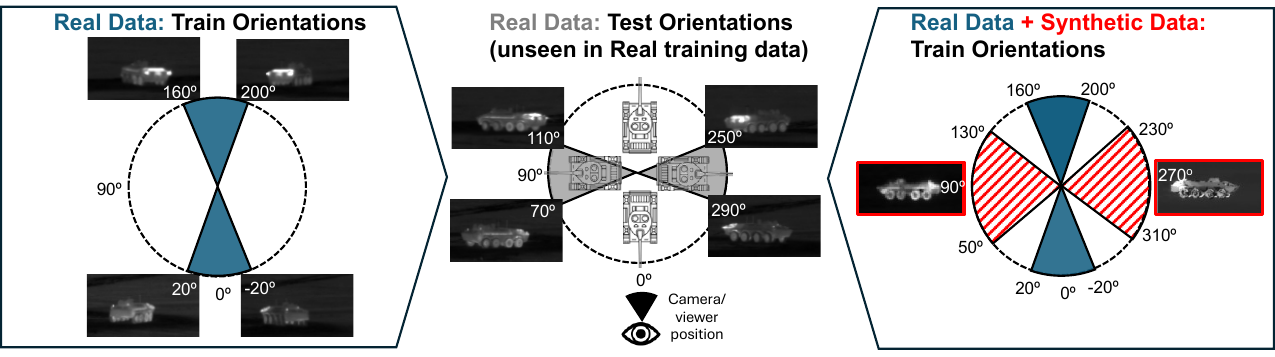}
    \caption{Experimental Setup. (left) Training data for only real data, showing vehicle orientations (blue segments) for all classes in training dataset from DSIAC ATR dataset. (middle) Testing data, showing vehicle orientations (gray segments) for all classes in test dataset. (right) Training data for real + synthetic data, showing vehicle orientations for real data (blue segments) and for synthetic data (red striped segments). Example cropped images shown for BTR70 class for orientations given in white text.}
    \label{fig:train-test-data}
\end{figure*}


Our aim is to perform object detection, that is, to detect and classify any vehicle(s) in the image, and estimate the bounding box coordinates of any detected vehicles. 
We use a YOLOv8 nano (You Only Look Once) model by Ultralytics \cite{yolov8_ultralytics} that has been pretrained on the COCO dataset, and fine-tune it on the ATR DSIAC dataset.



In order to prepare the dataset, we first crop the input images of initial size height 512, width 640 to a size of $256$ by $512$ pixels, with the centre of the crop positioned at a random pixel in the input image, which also provides negative samples where no vehicle is present. The cropped image is then resampled using bilinear interpolation to a size of $128$ by $256$ pixels. We split the images according to vehicle orientations. For our test and real training datasets, we use ATR DSIAC images with vehicle orientations in the ranges $\mathcal{O}_{test}=[70^\circ,110^\circ] \cup [250^\circ,290^\circ]$ and $\mathcal{O}_{train}=[-20^\circ,20^\circ]\cup [160^\circ,200^\circ]$ respectively. For synthetic training data, vehicle orientations are in the range $\mathcal{O}_{syn}=[50^\circ,130^\circ] \cup [230^\circ,310^\circ]$, as illustrated in Fig. \ref{fig:train-test-data}.
We employ this train-test split so that we can examine how the addition and modification of synthetic data improve our model's ability to detect orientations of target vehicles that have not been seen in training; synthetic data have especial utility for detection of unseen scenarios. This split also reduces data leakage between the train and test data, which can occur if train and test frames are sampled uniformly from the video dataset, as per much of the previous literature on the ATR DSIAC dataset \cite{Priddy2023ExplorationsIT, 2019SPIE10988E..08M, Sami2023DeepTT, 2024SPIE13083E..17O, 9438143}. As neighbouring frames in a video are highly correlated, results reported on such uniformly sampled train-test splits can be unrealistically high due to overfitting. 


\section{Results}
\label{sec:results}

To provide a baseline reference for improvements in performance we first fine-tune our YOLOv8n model using only the real data from the dataset, trained using $9000$ images from $\mathcal{O}_{train}$ and tested using $\mathcal{O}_{test}$, where there is no overlap between the vehicle orientations in $\mathcal{O}_{train}$ and $\mathcal{O}_{test}$. We report performance using the mean average precision score at an intersection over union (IoU) threshold of $0.5$ (mAP50). We take ensemble averages over 4 different random seeds to demonstrate that our model improvements are robust to seed variations. The baseline performance for YOLOv8n when the model is only trained on the real dataset shows an mAP50 score of $90.0\%$ (Table \ref{tab:exp2}, first row). This increases to $94.6\%$ when we instead train the YOLOv8n model on the same real data plus 9000 synthetic images generated from our initial mesh models, which we term v0 (version 0) and which includes vehicle orientations ($\mathcal{O}_{syn}$) that overlap the test time orientations in the real data $\mathcal{O}_{test}$ (compare Fig. \ref{fig:train-test-data} centre and right panels). Through a suite of tests we show that this increase in performance is not simply due to increased dataset size, and that the optimal ratio of real to synthetic data is 0.5 (supplementary Fig. 17), which we adopt through the rest of our experiments.


\begin{table*}[h]
    \centering
  \begin{tabular}{l|c | c | c}
    Training Data & \multicolumn{3}{c}{mAP50} \\
    \hline 
      & YOLOv8n & YOLOv8s & YOLOv8x\\
    \hline 
    Real & 90.0\% & - & - \\ 
    Real + Syn v0 (Initial) & 94.6\% & 95.6\% & 89.6\% \\
    Real + Syn v0 (Varifocal Loss) & 94.5\%  & - & - \\
    Real + Syn v0 (Online Hard Example Mining) & 94.4\% & - & -\\
    Real + Syn vR (Reinforcing)  & 95.7\% & 96.4\% & 91.6\%\\
    Real + Syn vD (Disruptive) & 95.7\% & 95.8\% & 89.9\%\\
    Real + Syn v(R+D) (Reinforcing and Disruptive) & 96.1\% & 96.4\% & 90.7\% \\
\end{tabular}
\caption{Performance improvement obtained from different synthetic data modifications measured in average mAP50 scores, where the average is taken over 4 different random seeds. The top row indicates the model trained only on real data. For the reinforcing modification unique features of the SUV are modified to reduce confusion between the SUV and BTR70. The disruptive modification refers to modifying the common features of the ZSU23 to reduce confusion with the BTR70.}
\label{tab:exp2}
\end{table*}

\subsection{Using XAI to improve model performance}
\label{sec:XAI_examples}

\begin{figure}[ht]
    \centering
    \includegraphics[width=0.48\textwidth]{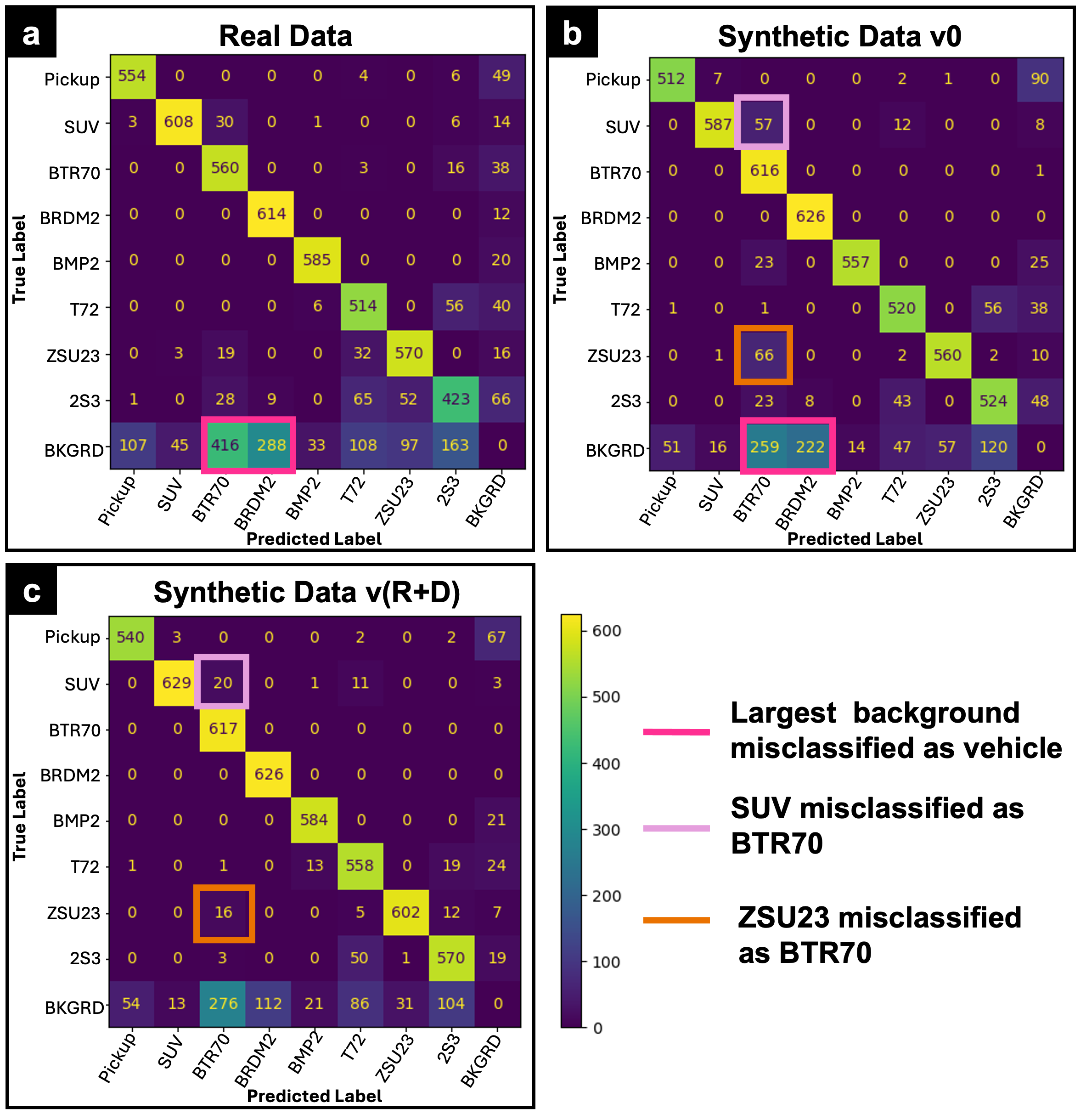}
    \caption{Confusion matrices calculated by taking an average over $4$ different seeds for models trained with real data plus (a) no synthetic data (b) synthetic data v0, (c)  synthetic data v(R+D) which includes both reinforcing modification, vR, to unique features of SUV relative to BTR70 and disruptive modification, vD, to common features of ZSU23 and BTR70.}
    \label{fig:AVG_CMS}
\end{figure}
\begin{figure*}[h]
\centering
\includegraphics[width=0.8\textwidth]{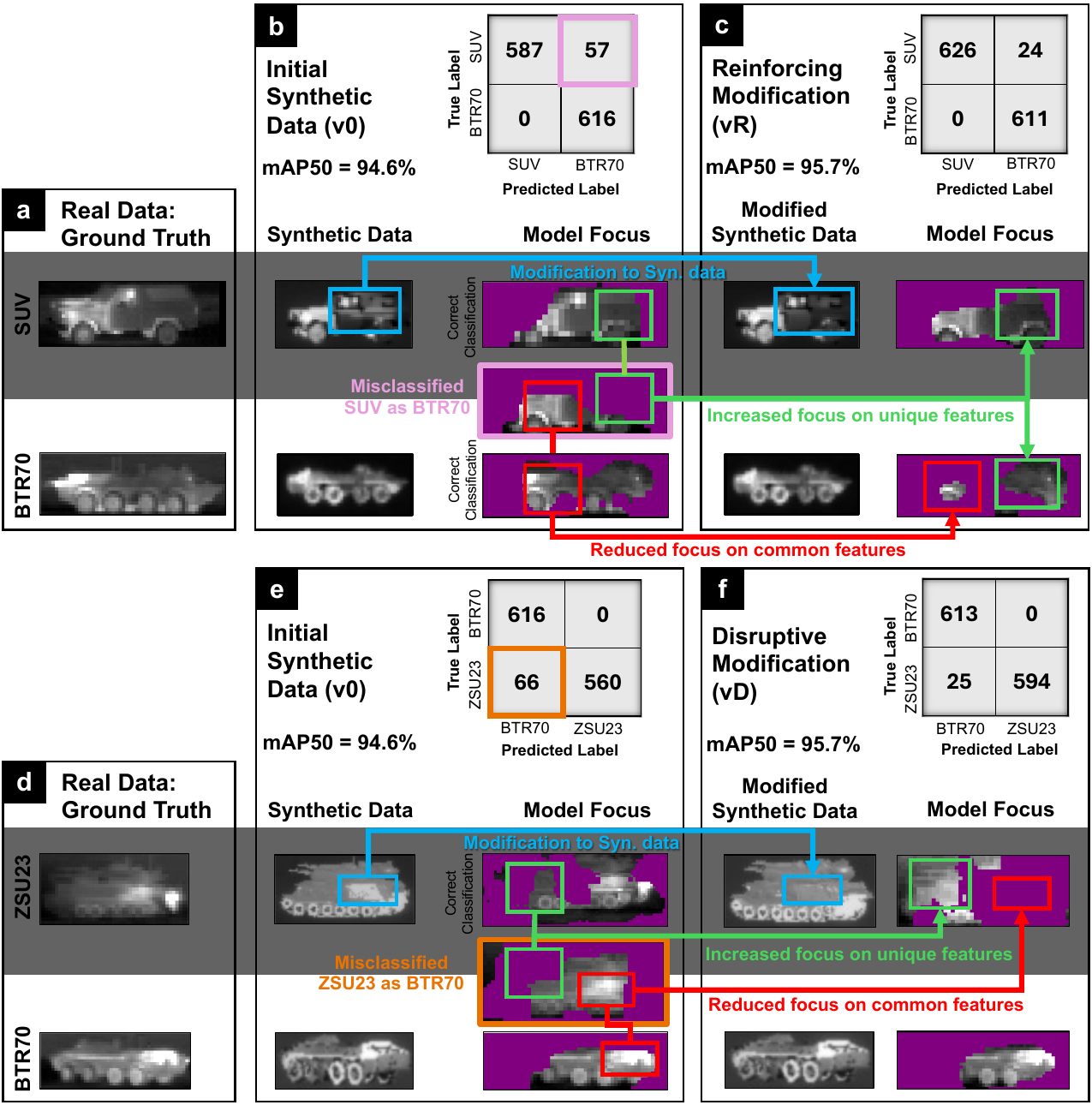}
    \caption{Illustration of the synthetic data modification process for reducing confusion. \textbf{a,d:} Ground-truth samples. \textbf{b,e:} Top shows confusion matrix for model trained on the initial synthetic data showing misclassifications highlighted in the confusion matrix. Bottom shows synthetic data samples (left), and SHAP contribution plots (right) for correct classifications capturing the unique features, and for misclassifications that capture the common features. \textbf{c,f:} Top shows confusion matrix for model trained on modified synthetic data, showing reduced misclassifications (see Supplementary Fig. 3 for full confusion matrices). Bottom shows samples of modified synthetic data (left), and SHAP contribution plots (right) for correct classifications illustrating an increased focus on unique features and reduced focus on common features. Pixels with a contribution score $<50\%$ (chosen to represent a trade-off between information and noise) of the highest score are masked in purple.}
    \label{fig:shap}
\end{figure*}

To understand the increase in mAP50 score from the addition of the v0 synthetic data to the training dataset, we compare the confusion matrices in Fig. \ref{fig:AVG_CMS}a,b and see a large reduction in the background being misclassified as vehicles (e.g. dark pink box). By plotting the SHAP values of the pixels, averaged over $100$ random correctly classified test images, we see an increase in focus inside the bounding box of the target vehicle (average SHAP values increase from $39\%$ to $54\%$, see supplementary Fig. 2), which indicates the reduction in misclassification of the background is due to increased model focus on the vehicles relative to the background, when synthetic v0 data are added in training. 
However, with the addition of synthetic data v0, the confusion matrices also show a small increase in confusion (decrease in performance) between some vehicles in the dataset e.g. BTR70 misclassified as SUV or ZSU23 (light pink and orange boxes respectively in Fig. \ref{fig:AVG_CMS}b and Fig. \ref{fig:shap}b,e). In the following sections we target these misclassifications using our new procedure from Section \ref{sec:procedure}, demonstrating the use of both Reinforcing and Disruptive modifications to the synthetic data v0 in order to reduce model confusion and improve performance.



\noindent{}\textbf{Reinforcing Modification Example:} Fig. \ref{fig:shap}a-c shows how we use Reinforcing Modification to reduce misclassification of the SUV class as a BTR70 (57 instances, light pink box in cropped and full confusion matrix in Fig. \ref{fig:shap}b and Fig. \ref{fig:AVG_CMS}b). 
To identify the cause of confusion we group the validation images into $5^{\circ}$ segments of vehicle orientations, which are this dataset's natural dimension of variation, and plot three average SHAP maps within each cluster; one each for the correct classification of the SUV and BTR70 and one for the misclassification of the SUV as a BTR70. 
Correct classifications of the SUV have increased focus towards the rear of the vehicle (Fig. \ref{fig:shap}b, top green box), whereas misclassifications of the SUV as a BTR70 do not focus in this same area (Fig. \ref{fig:shap}b, bottom green box) and instead have increased focus on the front bonnet of the SUV, which is a brighter (hotter) area in the IR images. This bright area is in a similar location to the hot rear exhaust of the BTR70 when the SUV is orientated $70^{\circ}-75^{\circ}$ (see Supplementary Fig. 18a,b) and the BTR70 is orientated $260^{\circ}-265^{\circ}$ (Fig. \ref{fig:shap}b, red boxes, Supplementary Fig. 18c,d). The bright SUV bonnet and BTR70 exhaust therefore represent common features and the side and rear of the SUV represent unique features. 
We \textit{reinforce} model focus on the unique features of the SUV by making the main frame and rear part of the synthetic SUV images more similar to the real SUV images, by reducing the smoothness parameter of the material on the SUV mesh model's surface. This reduces reflections off the smooth SUV surface (visible as bright-to-dark gradients in Fig. \ref{fig:shap}b, blue box) and results in more uniform texture (Fig. \ref{fig:shap}c, blue box), matching the real image (Fig. \ref{fig:shap}a, top).
The effect of this modification is to increase the focus on the unique features and accordingly reduce the focus on common features (Fig. \ref{fig:shap}c). This results in a reduction of SUV, BTR70 misclassifications from 57 to 24 and an mAP50 increase of 1.1\% from 94.6\% to 95.7\% (Fig. \ref{fig:shap}b,c, Table \ref{tab:exp2}). 




\noindent{}\textbf{Disruptive Modification Example:} 
Fig. \ref{fig:shap}d-f shows how we use Disruptive Modification to reduce misclassification of the ZSU23 class as a BTR70 (66 instances, see Fig. \ref{fig:AVG_CMS}b,c). In this case we note that the bright spot corresponding to the hot engine panel at the back of the ZSU23 in the orientation $105^{\circ}-110^{\circ}$ (see Supplementary Fig. 18a,b) is being confused with the hot rear exhaust of the BTR70 in the orientation $110^{\circ}-120^{\circ}$ (see Supplementary Fig. 18c,d). To disrupt the prominence of these common features we make the hot and bright engine on the rear of the ZSU23 significantly darker, in contrast to the real samples, reducing the realism of the synthetic data. This disruptive modification results in an increase in mAP50 of 1.1\% from 94.6\% to 95.7\% (Table \ref{tab:exp2}), similar to that achieved from using the reinforcing modification above. We also note that both types of modification result in better mAP50 scores than from use of varifocal loss \cite{9578034} or online hard example mining (OHEM, \cite{Shrivastava2016TrainingRO}) instead when training on the Real+Syn v0 dataset. These methods aim to improve classification performance of ``hard examples'' through optimising the training procedure, but these methods, used here as benchmarks, actually show a slight performance drop of 0.1-0.2\% compared to baseline results for the model trained on the Real+Syn v0 dataset (Table \ref{tab:exp2}, rows 2-6).

\noindent{}\textbf{Reinforcing + Disruptive:} When we combine both modifications, we observe reduction in the misclassification of BTR70 with both SUV and ZSU23 (Fig. \ref{fig:AVG_CMS}e), and a corresponding increase in mAP50 of 1.5\% on average from 94.6\% to 96.1\%; larger than the 1.1\% increase in performance gained from either modification alone (Table \ref{tab:exp2}).

These results are similar when we repeat our experiments with YOLOv8s and YOLOv8x architectures; the same modifications to the training data all lead to improved mAP50 scores averaged across 4 random seeds (Table \ref{tab:exp2}, final two columns). For these larger models, the reinforcing modification results in a larger mAP50 increase than the disruptive modification, but some variation is expected as for simplicity we have used the same modifications to the synthetic data that we derived from the YOLOv8n model - in reality we expect the exact nature of modifications required to vary with model architecture. 

%% file: sec_iccv/4_discussion.tex
\section{Discussion}
\label{sec:discussion}


We have demonstrated that XAI techniques can be used to guide modification of synthetic datasets used to train object detection models, in order to improve model performance. Importantly, our methodology reduces the workload of human operators in designing and optimising such datasets, and also allows both modifications that increase and decrease realism in synthetic data, through reinforcing unique features that distinguish between classes and through disrupting common features between classes. We show that both types of modifications can be used to reduce misclassifications and improve model performance. As a first proof of concept this demonstrates the power of XAI-driven modifications to generate better synthetic data more efficiently. 


Our current results already have potential to enable faster generation of synthetic datasets in a human-in-the-loop process. But although such human-in-the-loop processes are often desirable for maintaining control over trained deep learning models, future extension of this work could move towards partial automation. This could be through automated modification of the 3D mesh model by mapping the 2D SHAP images to the 3D mesh using ray-tracing \cite{2d-3d_raytrace}, modifying the mesh corresponding to unique or common features by changing the material properties adding or removing sub-parts of the vehicle, rotating or scaling sub-parts, and then training an additional model to learn which modifications lead to the largest increases in performance. Alternatively our method is not limited to use of game engines, and could also work with diffusion models if they can provide the level of fine-grained control needed for modifications.
Furthermore, while we currently use SHAP to show the location of common and unique features in the input image, mechanistic interpretability \cite{bereska2024mechanistic} could be used to also show the nature of these features (shape, texture, imperfections, etc.), which would allow us to more efficiently and effectively modify our mesh models.

Introducing synthetic samples for unseen scenarios and steering a model's attention in specific directions also presents a novel method to obtain more robust, unbiased, and safe models for critical applications such as autonomous driving \cite{10.1145/3702989} and medical diagnosis \cite{MIHAN2024e749}, which are sensitive to edge cases and class confusion, and have shown bias towards racial, age or gender characteristics \cite{10.1145/3702989}. Interpreting the location and nature of such a model's focus and procedurally generating appropriate synthetic samples to counter these effects could allow us to steer the model to instead focus on more relevant features, and thus improve its robustness and generalisation. 

Steering model decisions through careful, explainable AI-guided curation of synthetic data enables us to identify and fix bugs, flaws and systematic biases in computer vision models, whilst simultaneously overcoming a challenge of critical practical importance; reducing the workload of human operators in designing and optimising synthetic datasets. 

\noindent{}\textbf{Acknowledgement:} Research was sponsored by the Army and was accomplished under Cooperative Agreement Number W911NF-22-2-0161. The views and conclusions contained in this document are those of the authors and should not be interpreted as representing the official policies, either expressed or implied, of the U.S. Army or the U.S. Government. The U.S. Government is authorized to reproduce and distribute reprints for Government purposes notwithstanding any copyright notation herein.

%% file: sec_iccv/X_suppl.tex
\setcounter{page}{1}
\maketitlesupplementary

\section{Dataset}
\label{sec:supp_dataset}

The complete ATR DSIAC dataset \cite{DSIAC}, captured in 2006, contains 9 vehicle targets plus human targets, ground truth data including vehicle labels and bounding box coordinates, and other documentation including photographs of the targets and meteorological data to assist the user in correctly interpreting the imagery. The list of vehicles consists of: Pickup Truck; sport utility vehicle (SUV); BTR70 armored personnel carrier; BRDM2 infantry scout vehicle; BMP2 armored personnel carrier; T72 main battle tank; ZSU23-4 anti-aircraft weapon; 2S3 self-propelled howitzer; and MTLB armored reconnaissance vehicle. All imagery was captured using commercial cameras operating in the medium-wave infrared red (MWIR) and visible bands. 

\begin{figure}[h!]
    \centering
    \includegraphics[width=0.3\textwidth]{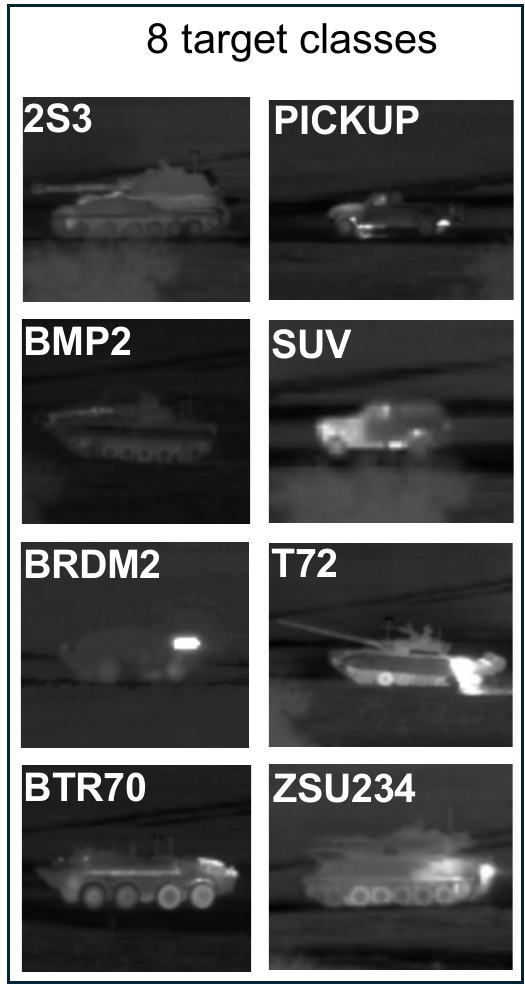}
    \caption{Example MWIR images for the 8 vehicles in the DSIAC ATR dataset.}
    \label{fig:vehicles}
\end{figure}

The dataset contains videos, containing $1800$ frames each, of targets at different distance ranges, including $1000$m, $1500$m, $2000$m, $2500$m, $3000$m, $3500$m, $4000$m, $4500$m, $5000$m, as well as in the day and at night. In this study we select a subset of ranges ($1000$m-$2000$m) to speed up model training, and because the vehicles in further distance ranges are low resolution due to the quality of the commercial cameras used in 2006, which makes identifying model modifications substantially more complex. We also only consider the 8 vehicle classes represented in Supplementary Fig. \ref{fig:vehicles}. The images featuring the MTLB and D20 always include both classes in a single image, and one vehicle often occludes the other. This introduces a unique variation from the other vehicle classes in the dataset, so we discard these 2 vehicle classes for this work. 

In the videos the targets move in approximately circular paths on the ground, and the camera is located at a slightly elevated position, so that a range of orientations (front, back, sides) are captured in the dataset. The videos in the dataset have stationary and similar backgrounds, and each video necessarily has successive frames that are almost identical. Therefore, the dataset lacks diversity of scenarios and scenes. Given this property, we take careful consideration when building a train and test set to avoid the issue of data leakage (see Supplementary Section \ref{sec:pca}).

\section{Background SHAP}

\begin{figure}[h!]
    \centering
    \includegraphics[width=0.45\textwidth]{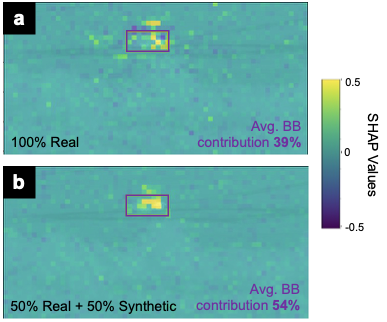}
    \caption{Example plots of SHAP contributions of pixels to outputs of models trained on (a) real data only and (b) real + synthetic v0 (initial) data, for a single test image of a ZSU23. In each image, the bounding box is shown in purple, and the average contributions for the two models over 100 random correctly classified test images such as this one, from within bounding box (BB), is given in the lower right corner. There is increased model focus on the vehicle for the model trained on both real and synthetic data (b), which is shown by the greater proportion of positive (yellow) SHAP values within the bounding box.}
    \label{fig:shap-background}
\end{figure}

In Section 5.1 in the main paper, we report and discuss the increase in focus of the model on features within the bounding box compared to outside the bounding box (i.e., background) when trained on real data supplemented with synthetic data. This result is illustrated in Fig. \ref{fig:shap-background} for a single example test image.

\section{Confusion Matrices}

\begin{figure*}[ht]
    \centering
    \includegraphics[width=0.8\textwidth]{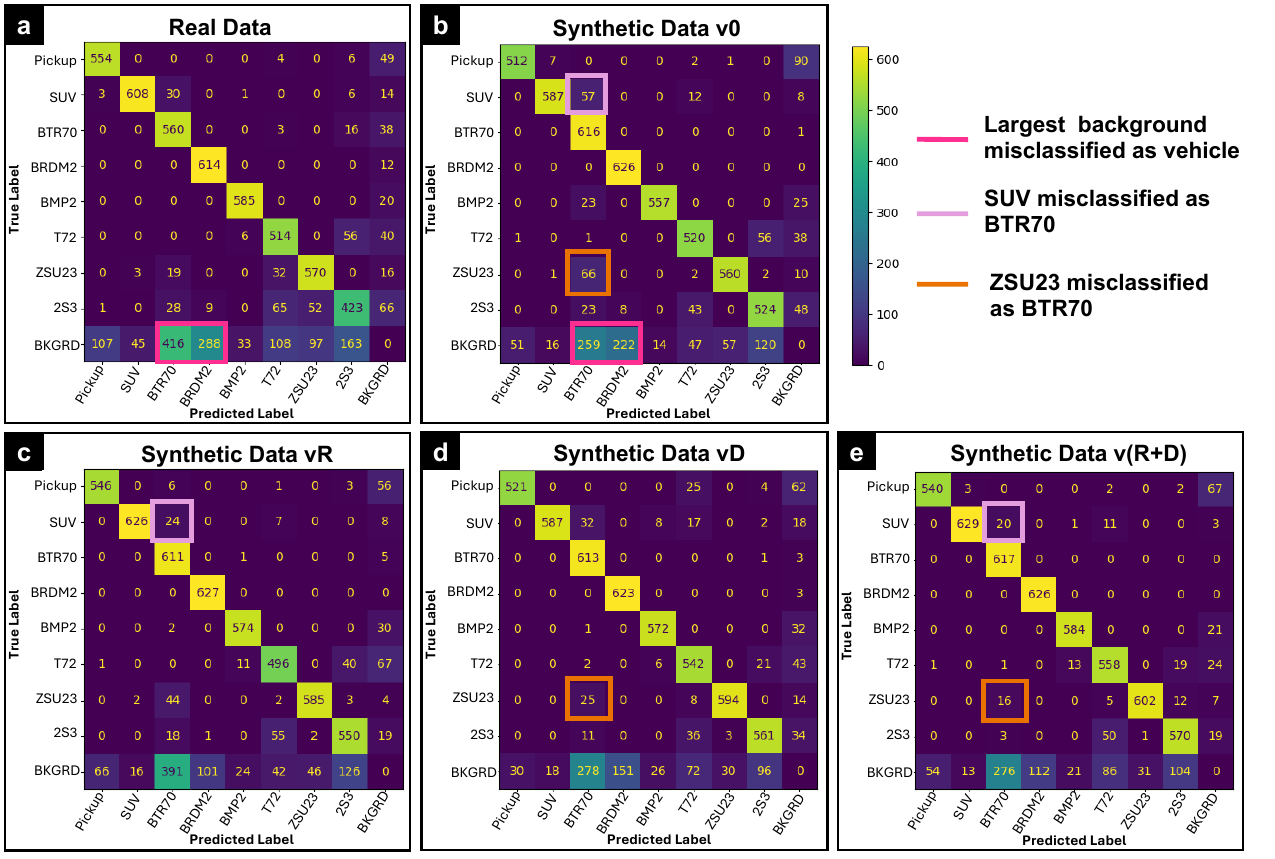}
    \caption{Confusion matrices calculated by taking an average over $4$ different seeds for YOLOv8n models trained with real data plus (a) no synthetic data (b) synthetic data v0, (c) synthetic data vR (reinforcing modification to unique features of SUV relative to BTR70), (d) synthetic data vD (disruptive modification to common features of ZSU23 and BTR70), and (e) synthetic data v(R+D)}
   \label{fig:AVG_CMS}
\end{figure*}

In Section 5 in the main paper, we present results of the ATR dataset trained on different synthetic data modifications. The full confusion matrices of the initial confusion when the model is trained on real data only and the real data and synthetic data constructed from the initial mesh models is reproduced from Fig. 4a,b in the main text in Fig. \ref{fig:AVG_CMS}a,b respectively. The remaining panels, c,d show the full confusion matrices corresponding to the truncated confusion matrices in Fig. 5c,f from the main paper for the reinforcing (vR) and disruptive (vD) modifications respectively. Again Fig. \ref{fig:AVG_CMS}e is a reproduction of Fig. 4c from the main paper corresponding to the confusion matrix for the reinforcing and disruptive v(R+D) modifications. 

\begin{figure*}[h!]
    \centering    \includegraphics[width=0.8\textwidth]{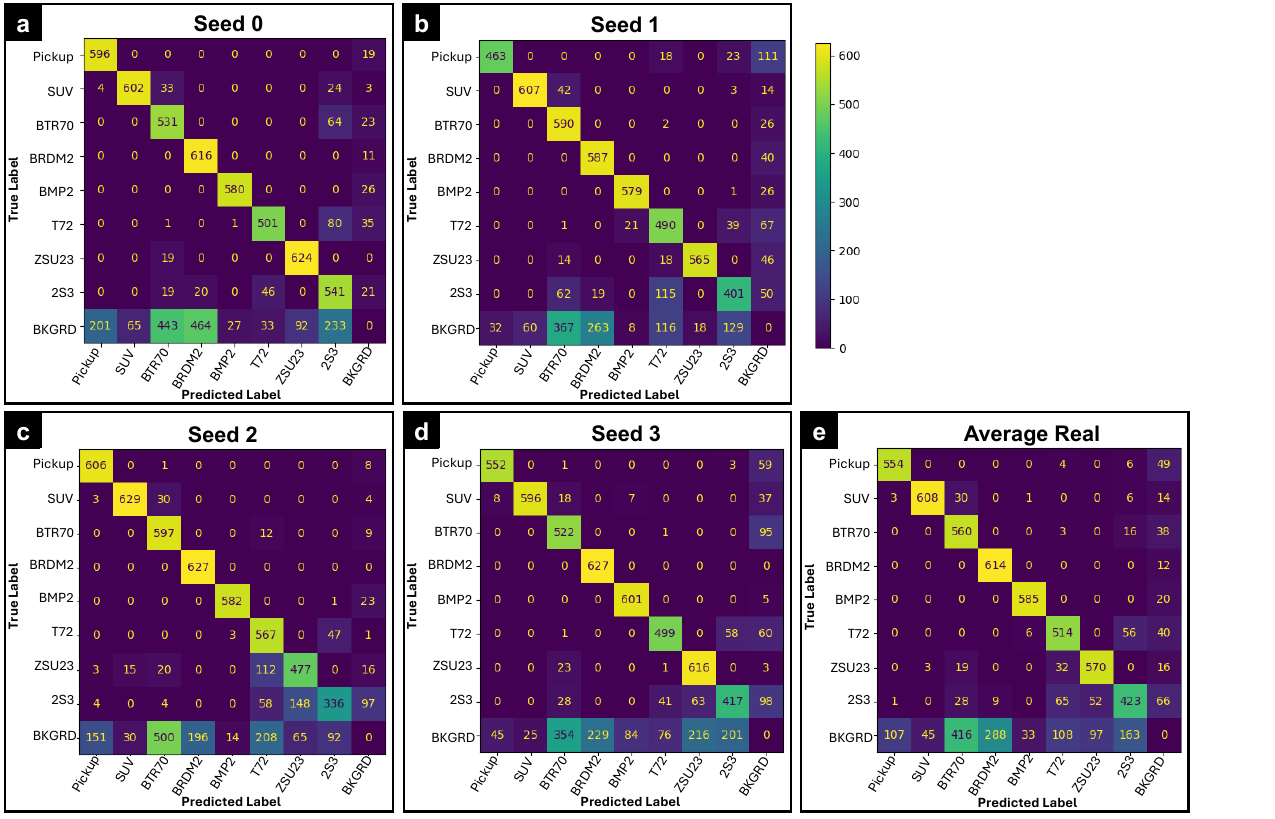}
    \caption{Confusion matrices for the YOLOv8n model trained on only real data from the dataset (i.e no synthetic data). Panels (a)-(d) correspond to 4 different seeds and panel (e) corresponds to the average over these 4 different seeds and is the same confusion matrix as that which appears in panel a of Fig. 4a in the main paper.}
    \label{fig:real_ensemble}
\end{figure*}

For each trained model in the main paper (reported in Table 1) we provided average confusion matrices (Fig. 4 in main paper). Here, in Supplementary Figs. \ref{fig:real_ensemble}-\ref{fig:mod1_and_mod2_ensemble}, we provide the complete set of four confusion matrices that were used to calculate this average, corresponding to the four random seeds used for each trained model. We also provide the confusion matrices for different seeds for experiments with YOLOv8s and YOLOv8x in Fig. \ref{fig:yolov8s_v0}-\ref{fig:yolov8x_vD}.

\begin{figure*}[h!]
    \centering
\includegraphics[width=0.8\textwidth]{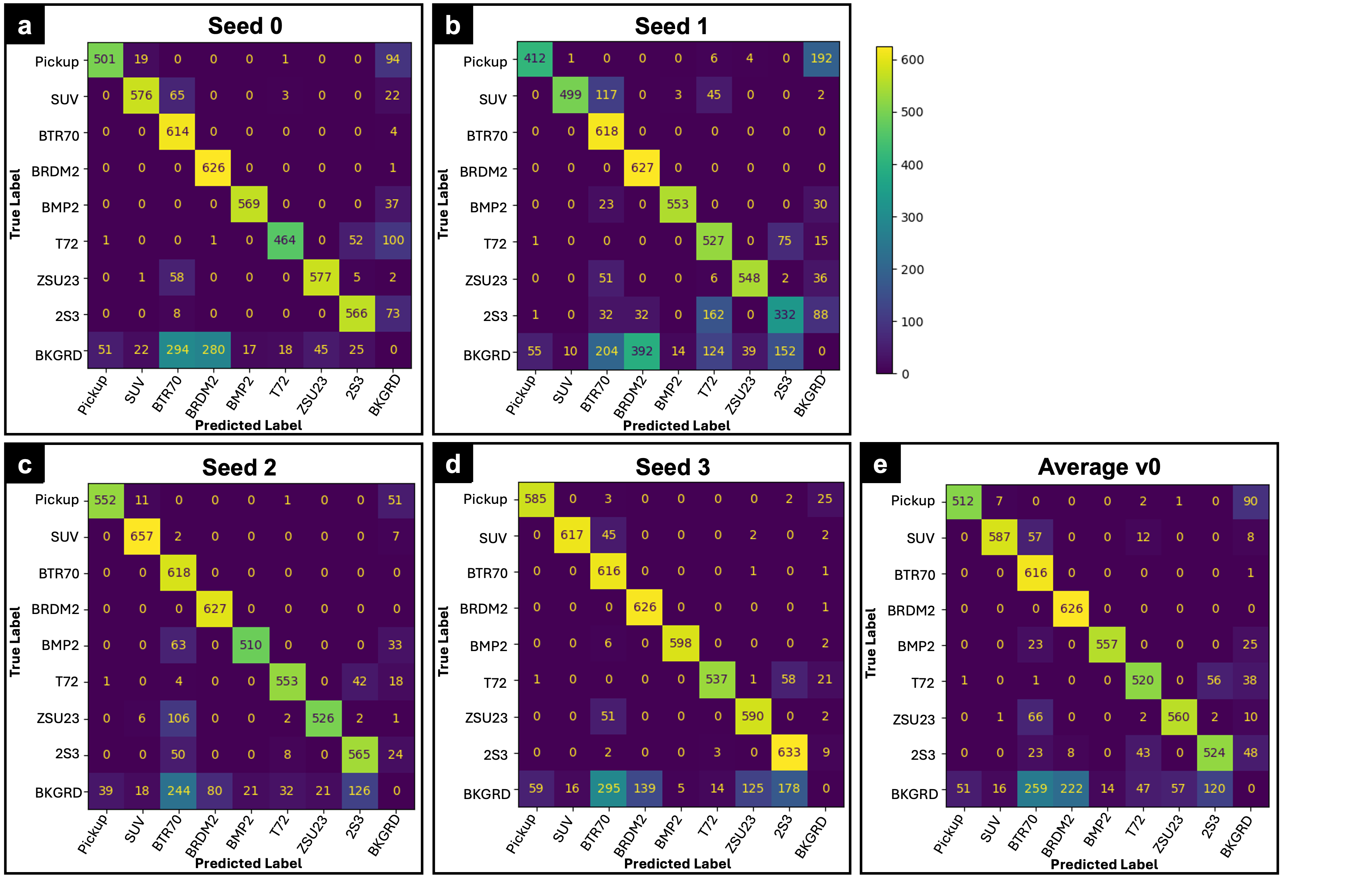}
    \caption{Confusion matrices for the YOLOv8n model trained on real data and the initial synthetic data (dataset v0). Panels (a)-(d) correspond to 4 different seeds and panel (e) corresponds to the average over these 4 different seeds and is the same confusion matrix as that which appears Fig. 4b in the main paper.}
    \label{fig:v0_ensemble}
\end{figure*}

\begin{figure*}[h!]
    \centering
    \includegraphics[width=0.8\textwidth]{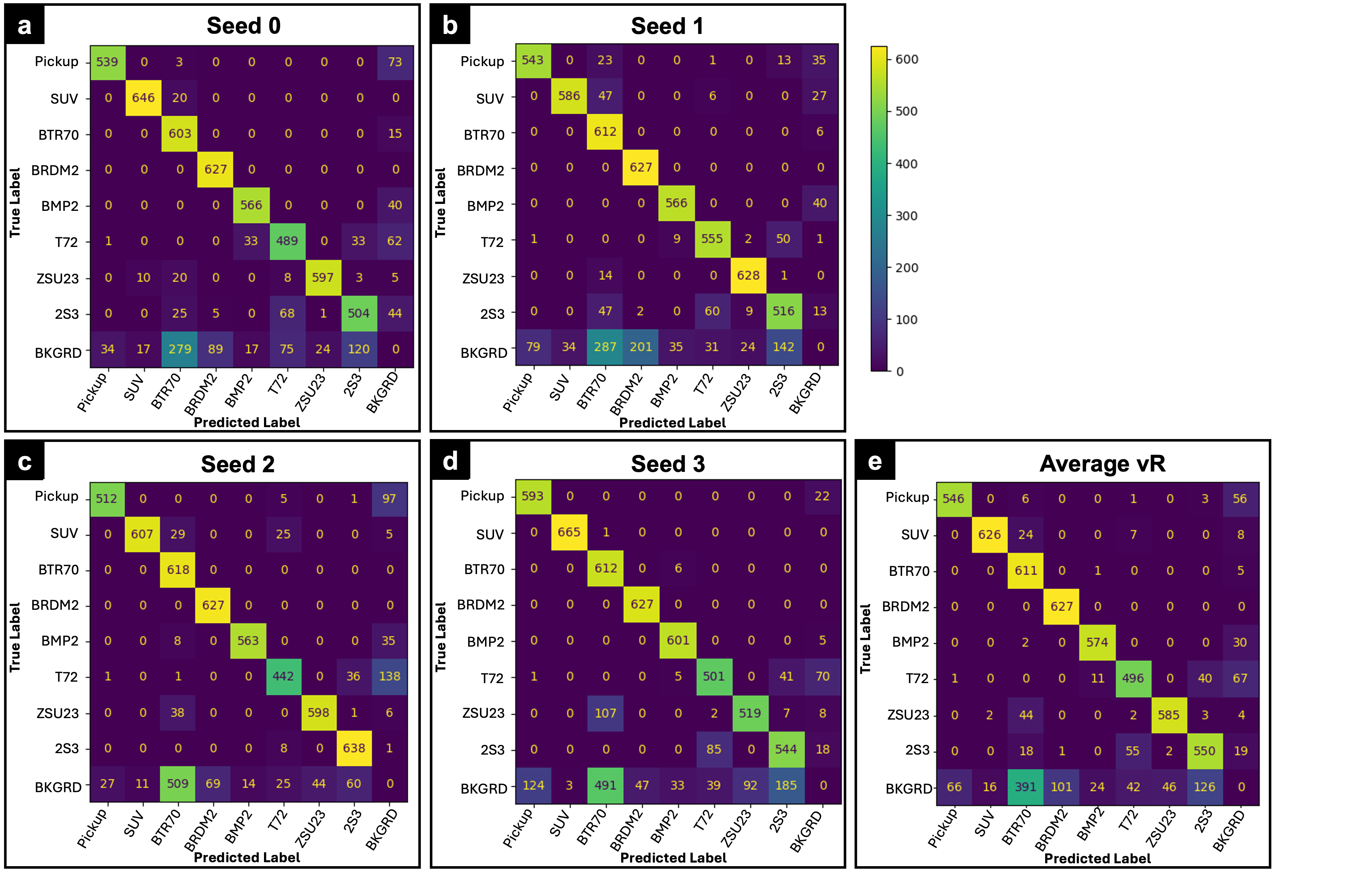}
    \caption{Confusion matrices for the YOLOv8n model trained on real data and synthetic data after the reinforcing modification (dataset vR) has been applied. Panels (a)-(d) correspond to 4 different seeds and panel (e) corresponds to the average over these 4 different seeds and is the same confusion matrix as that which appears in Fig. 4c in the main paper.}
    \label{fig:mod1_ensemble}
\end{figure*}

\begin{figure*}[h!]
    \centering
    \includegraphics[width=0.8\textwidth]{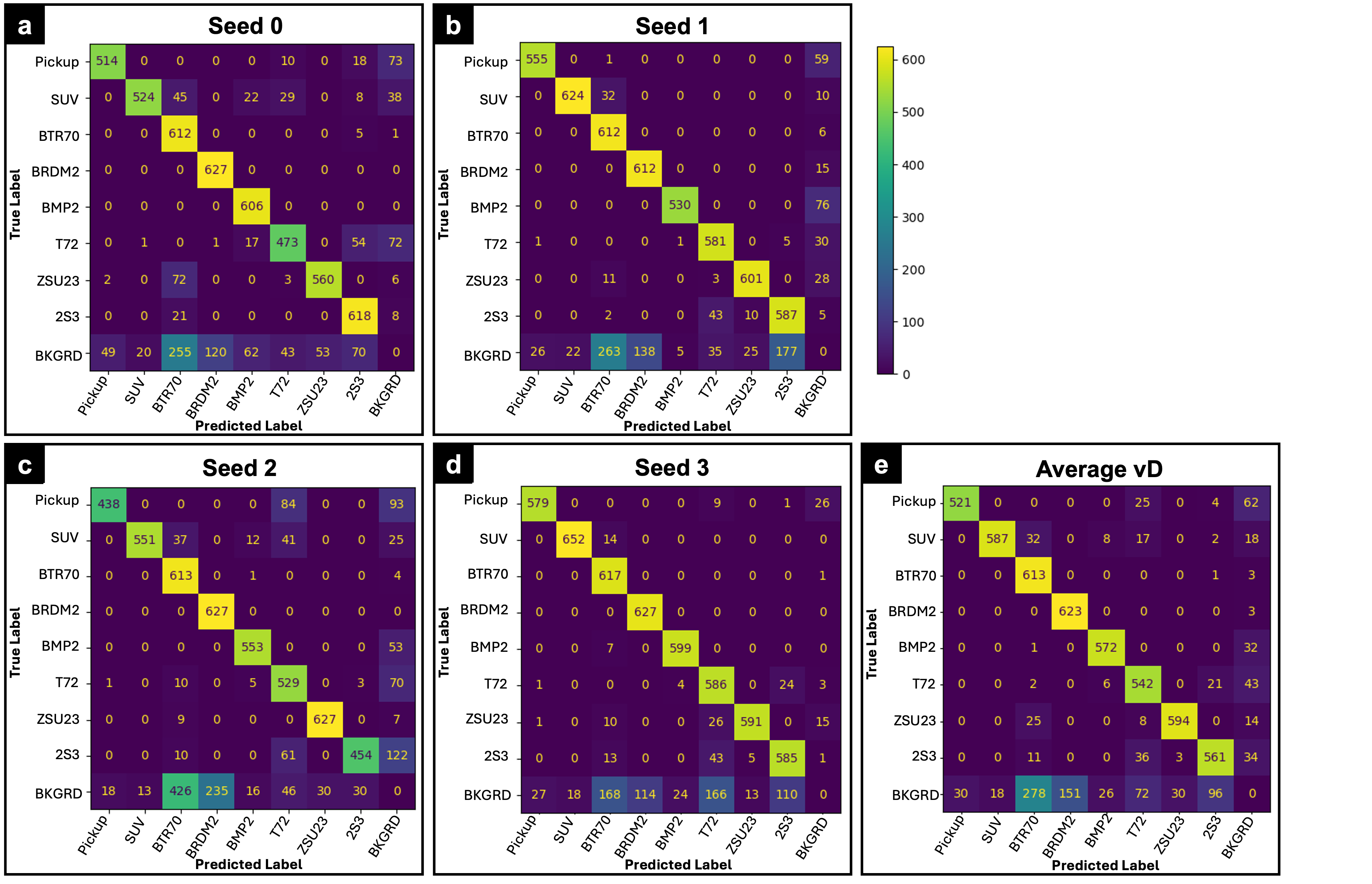}
    \caption{Confusion matrices for the YOLOv8n model trained on real data and synthetic data after the disruptive modification (dataset vD) has been applied. Panels (a)-(d) correspond to 4 different seeds and panel (e) corresponds to the average over these 4 different seeds and is the same confusion matrix as that which appears in Fig. 4d in the main paper.}
    \label{fig:mod2_ensemble}
\end{figure*}

\begin{figure*}[h!]
    \centering
    \includegraphics[width=0.8\textwidth]{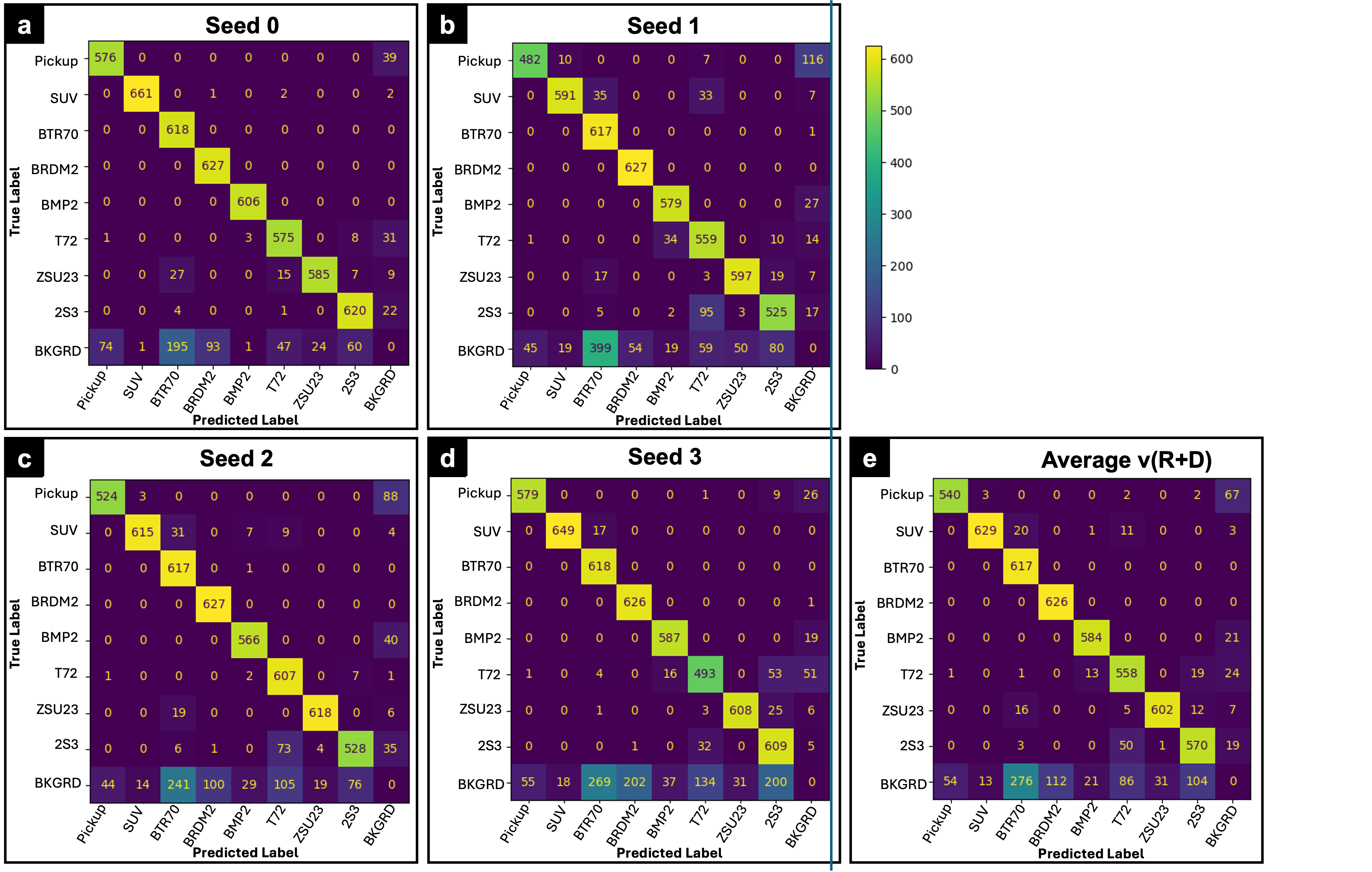}
    \caption{Confusion matrices for the YOLOv8n model trained on real data and synthetic data after the reinforcing and disruptive modifications (dataset v(R+D)) have been applied. Panels (a)-(d) correspond to 4 different seeds and panel (e) corresponds to the average over these 4 different seeds and is the same confusion matrix as that which appears in Fig. 4e in the main paper.}
    \label{fig:mod1_and_mod2_ensemble}
\end{figure*}

\begin{figure*}[h!]
    \centering
    \includegraphics[width=0.8\textwidth]{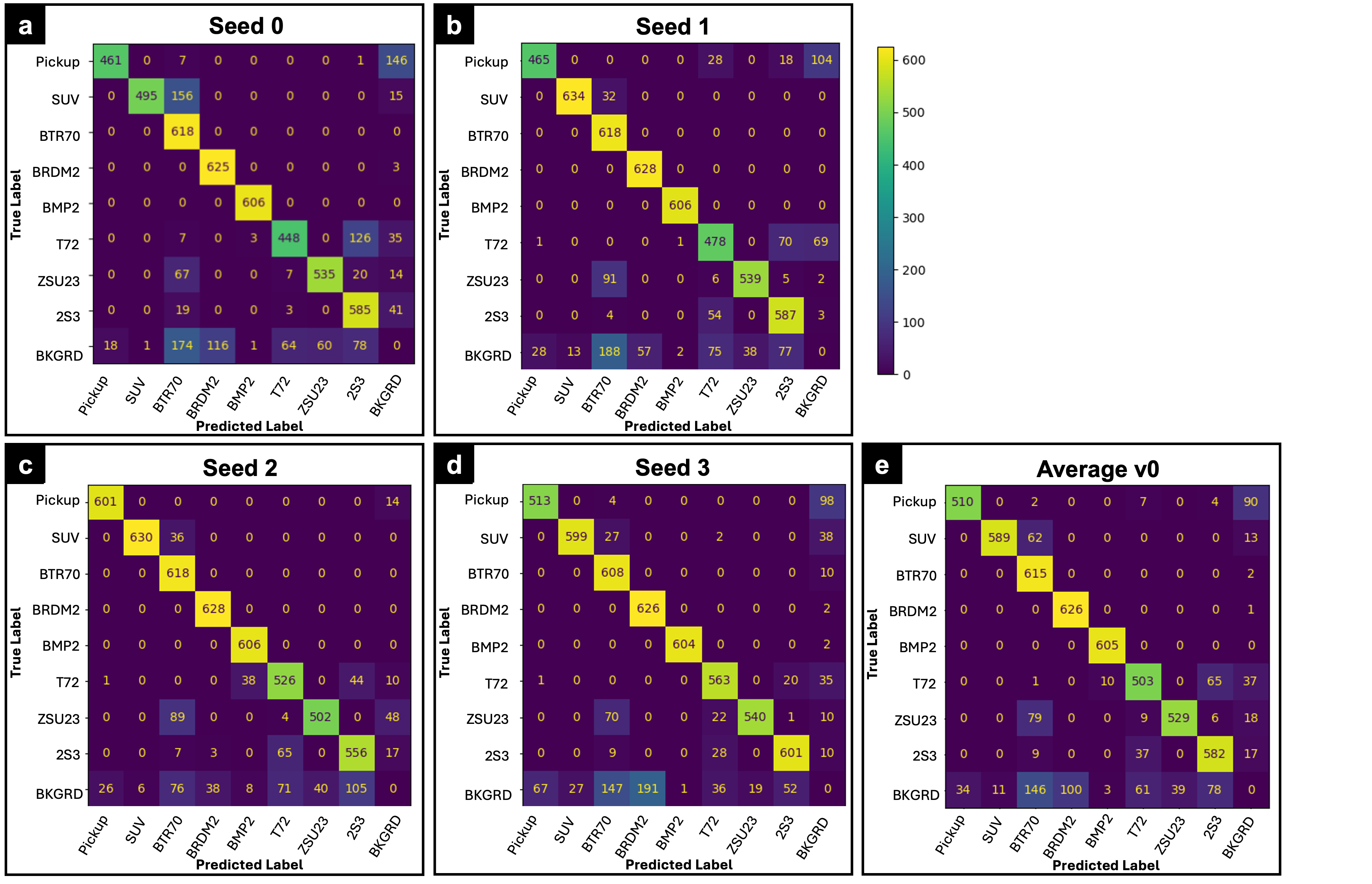}
    \caption{Confusion matrices for the YOLOv8s model trained on real data and initial synthetic data (dataset v0). Panels (a)-(d) correspond to 4 different seeds and panel (e) corresponds to the average over these 4 different seeds.}
    \label{fig:yolov8s_v0}
\end{figure*}

\begin{figure*}[h!]
    \centering
    \includegraphics[width=0.8\textwidth]{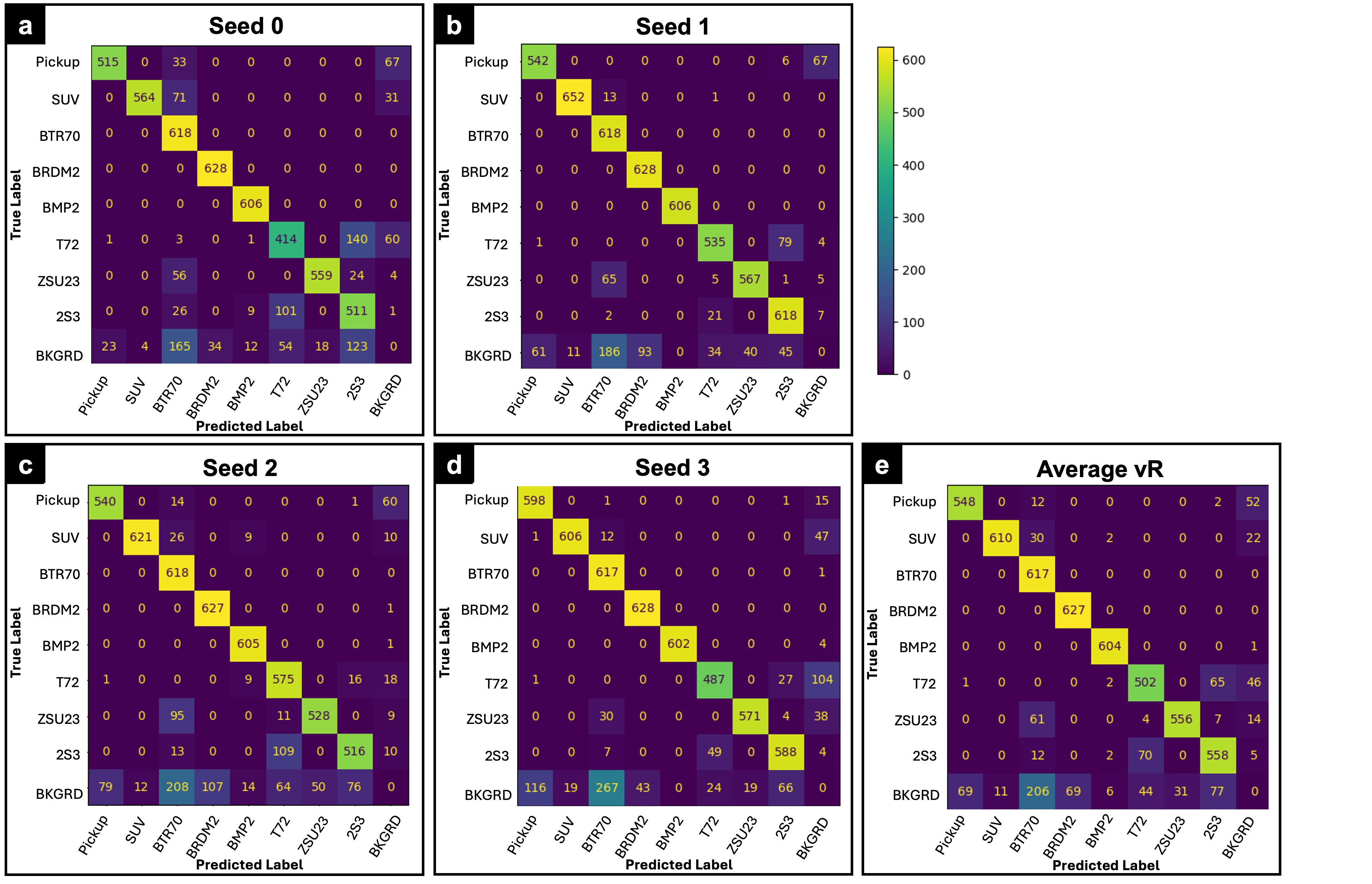}
    \caption{Confusion matrices for the YOLOv8s model trained on real data and synthetic data after reinforcing modification (dataset vR) has been applied. Panels (a)-(d) correspond to 4 different seeds and panel (e) corresponds to the average over these 4 different seeds.}
    \label{fig:yolov8s_vR}
\end{figure*}

\begin{figure*}[h!]
    \centering
    \includegraphics[width=0.8\textwidth]{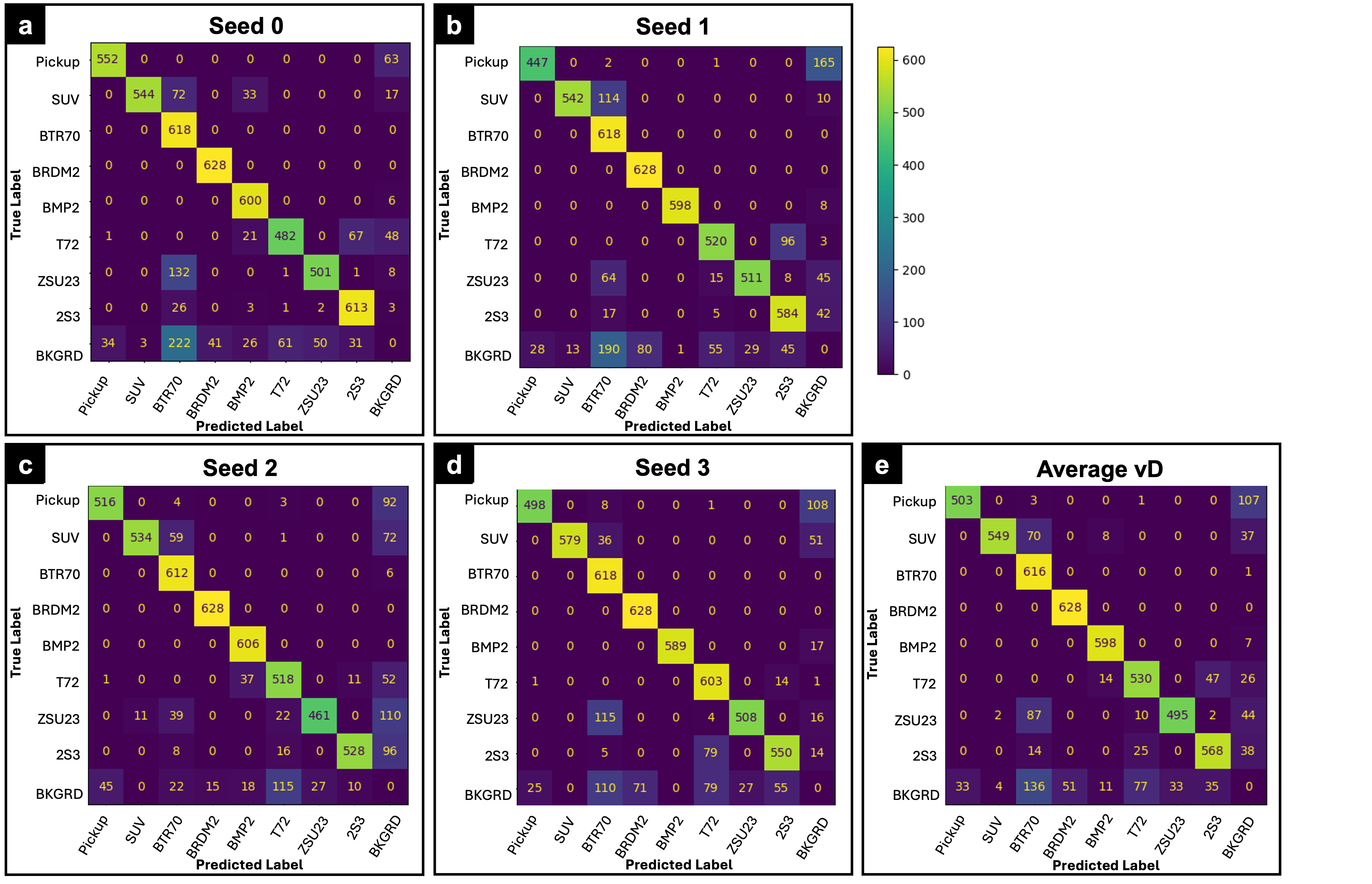}
    \caption{Confusion matrices for the YOLOv8s model trained on real data and synthetic data after disruptive modification (dataset vD) has been applied. Panels (a)-(d) correspond to 4 different seeds and panel (e) corresponds to the average over these 4 different seeds.}
    \label{fig:yolov8s_vD}
\end{figure*}

\begin{figure*}[h!]
    \centering
    \includegraphics[width=0.8\textwidth]{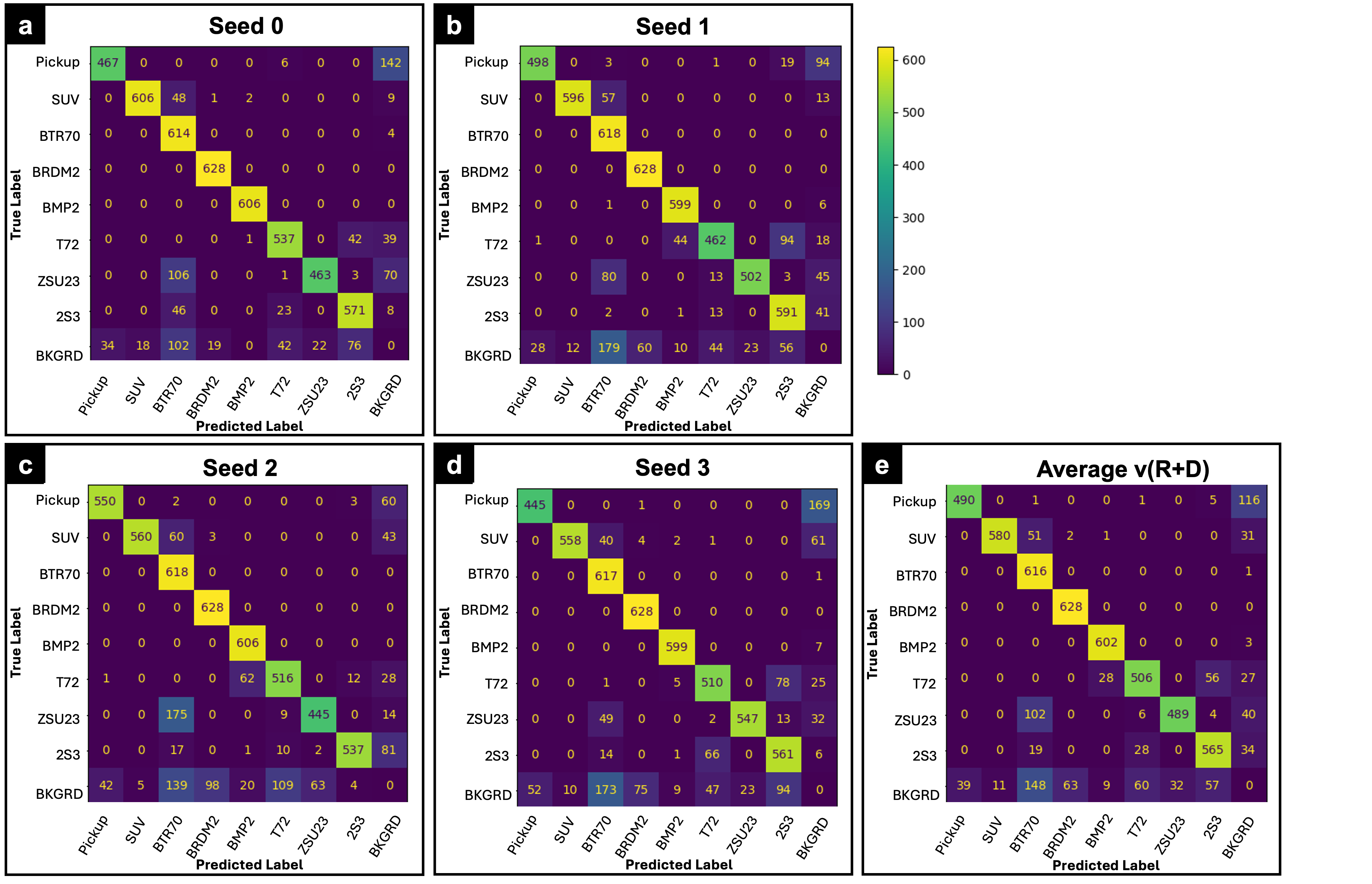}
    \caption{Confusion matrices for the YOLOv8s model trained on real data and synthetic data after reinforcing + disruptive modifications (dataset v(R+D)) have been applied. Panels (a)-(d) correspond to 4 different seeds and panel (e) corresponds to the average over these 4 different seeds.}
    \label{fig:yolov8s_v(R+D)}
\end{figure*}

\begin{figure*}[h!]
    \centering
    \includegraphics[width=0.6\textwidth]{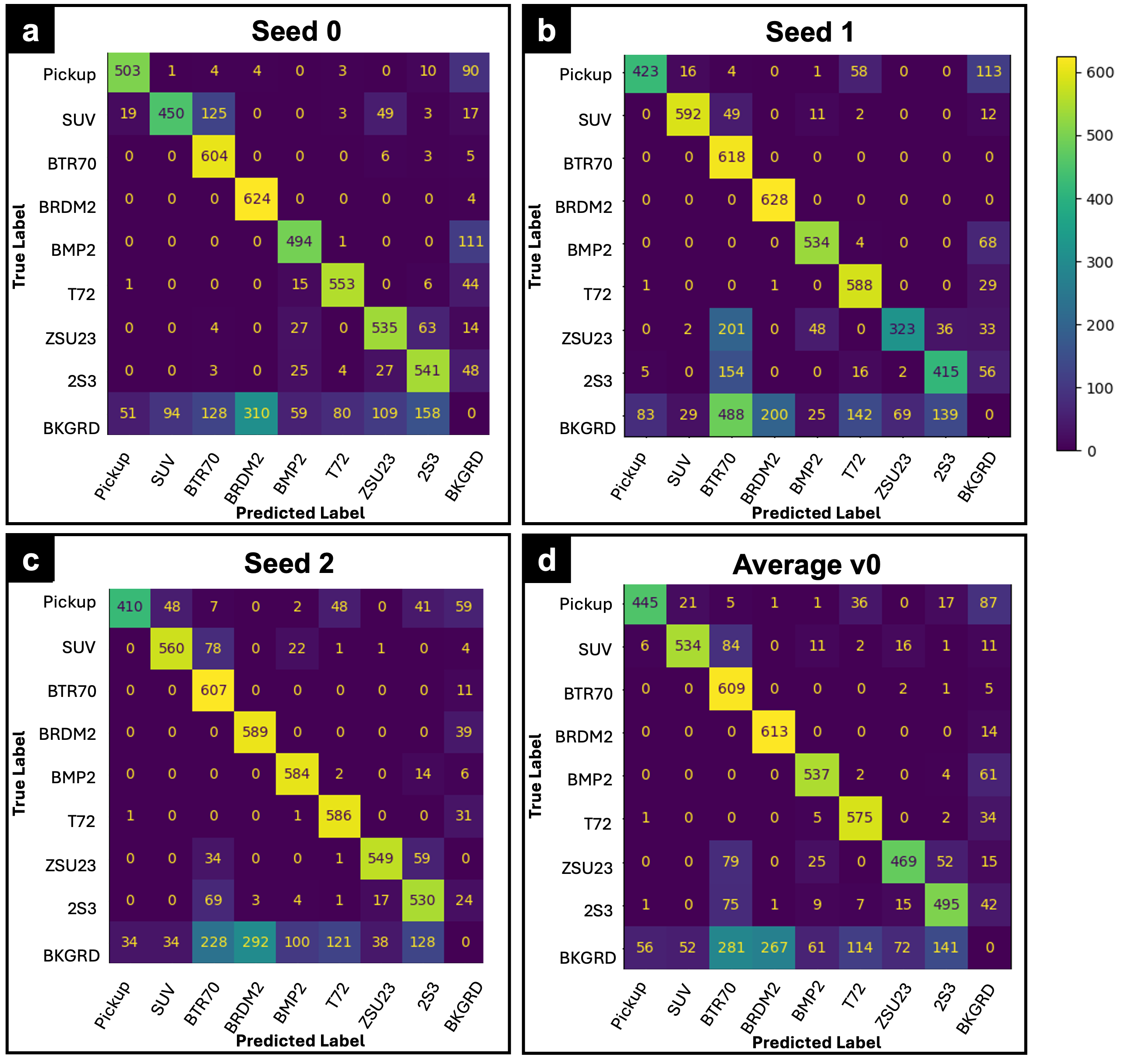}
    \caption{Confusion matrices for the YOLOv8x model trained on real data and initial synthetic data (dataset v0). Panels (a)-(c) correspond to 3 different seeds and panel (d) corresponds to the average over these 3 different seeds.}
    \label{fig:yolov8x_v0}
\end{figure*}

\begin{figure*}[h!]
    \centering
    \includegraphics[width=0.6\textwidth]{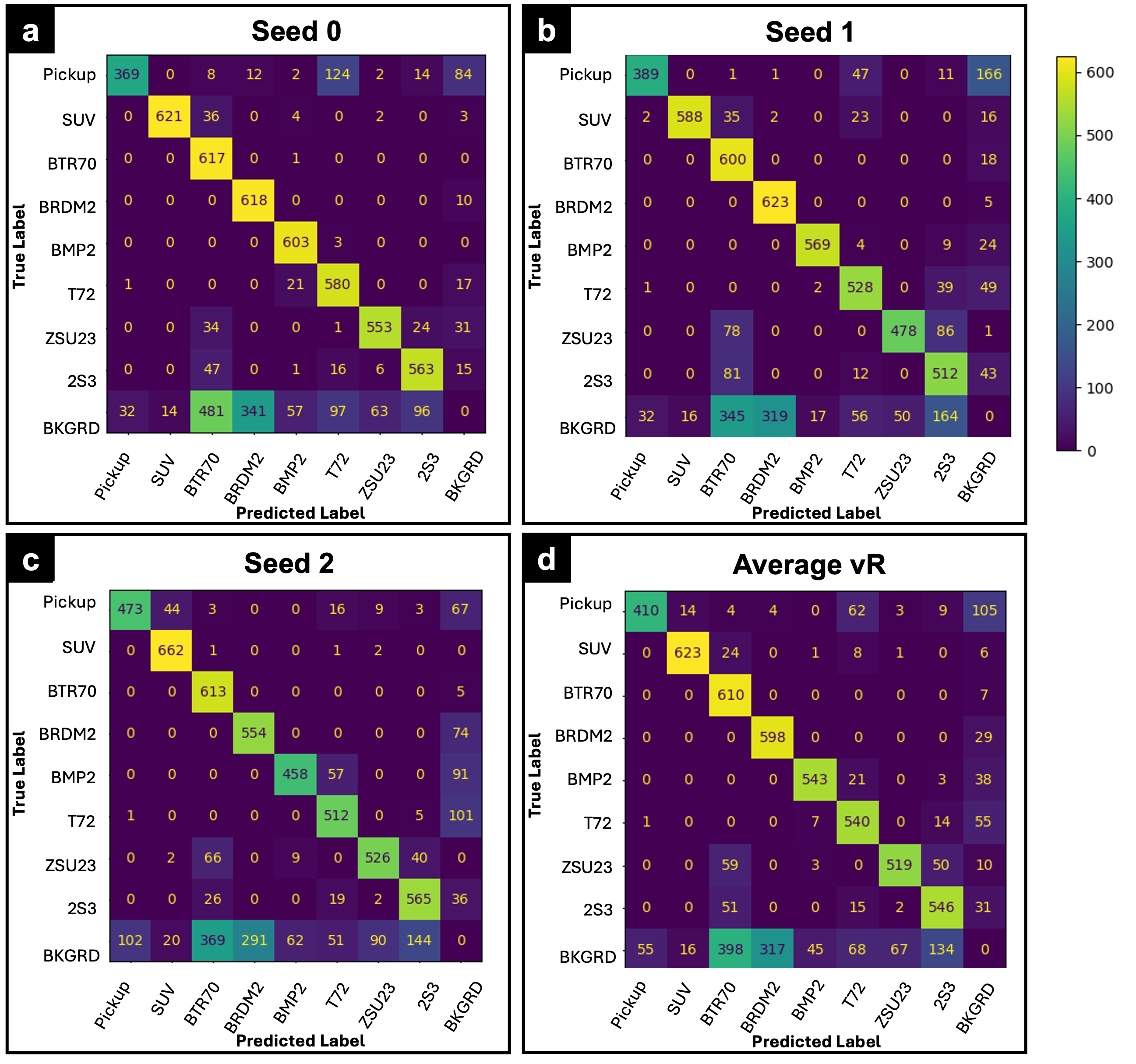}
    \caption{Confusion matrices for the YOLOv8x model trained on real data and synthetic data after reinforcing modification (dataset vR) has been applied. Panels (a)-(c) correspond to 3 different seeds and panel (d) corresponds to the average over these 3 different seeds.}
    \label{fig:yolov8x_vR}
\end{figure*}

\begin{figure*}[h!]
    \centering
    \includegraphics[width=0.6\textwidth]{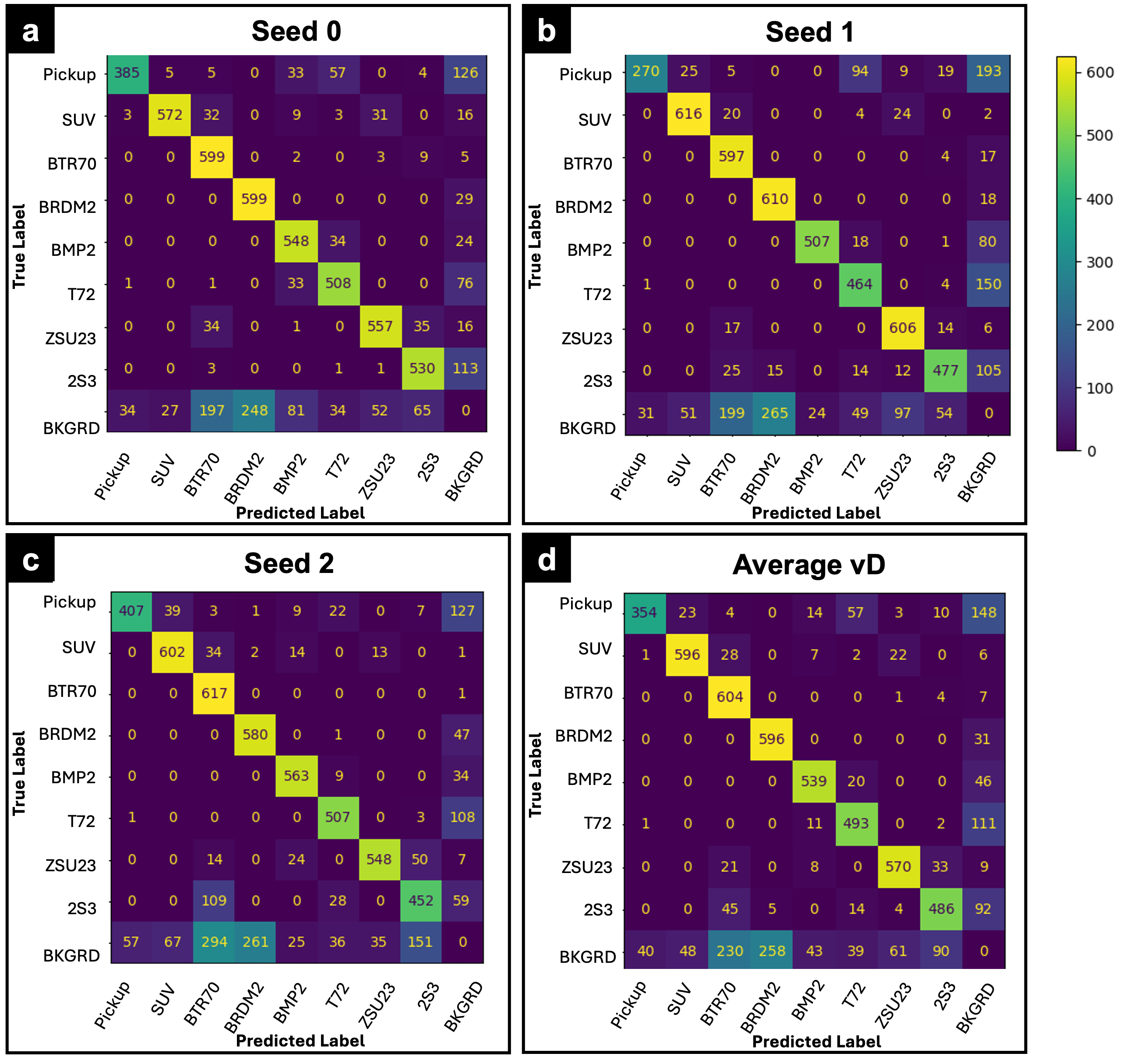}
    \caption{Confusion matrices for the YOLOv8x model trained on real data and synthetic data after disruptive modification (dataset vD) has been applied. Panels (a)-(c) correspond to 3 different seeds and panel (d) corresponds to the average over these 3 different seeds.}
    \label{fig:yolov8x_vD}
\end{figure*}

\begin{figure*}[h!]
    \centering
    \includegraphics[width=0.6\textwidth]{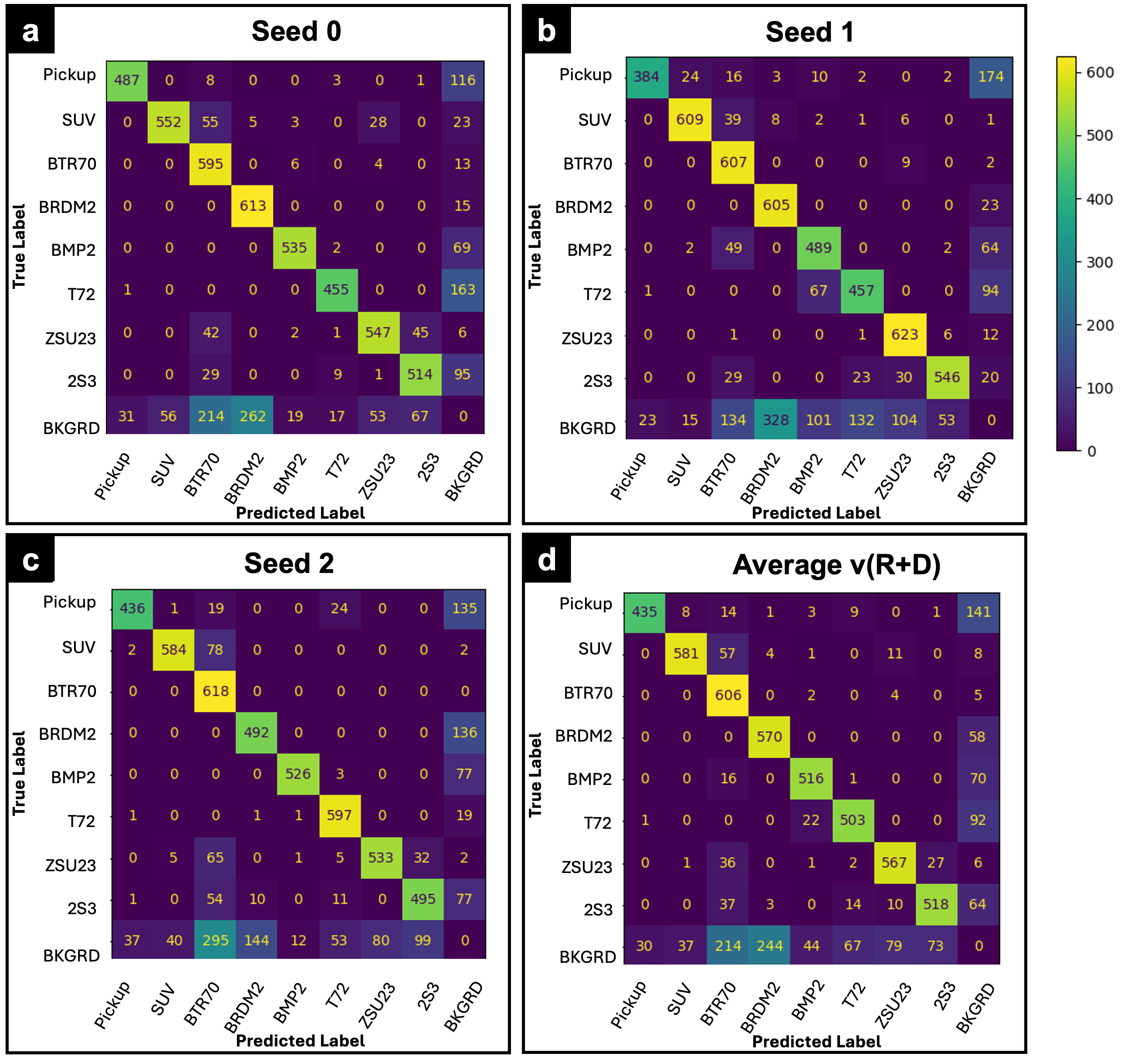}
    \caption{Confusion matrices for the YOLOv8x model trained on real data and synthetic data after reinforcing + disruptive modifications (dataset v(R+D)) have been applied. Panels (a)-(c) correspond to 3 different seeds and panel (d) corresponds to the average over these 3 different seeds.}
    \label{fig:yolov8x_v(R+D)}
\end{figure*}

\section{Model sensitivity to real:synthetic data ratio and dataset size}
\begin{figure}[ht]
    \centering
    \includegraphics[scale=0.45]{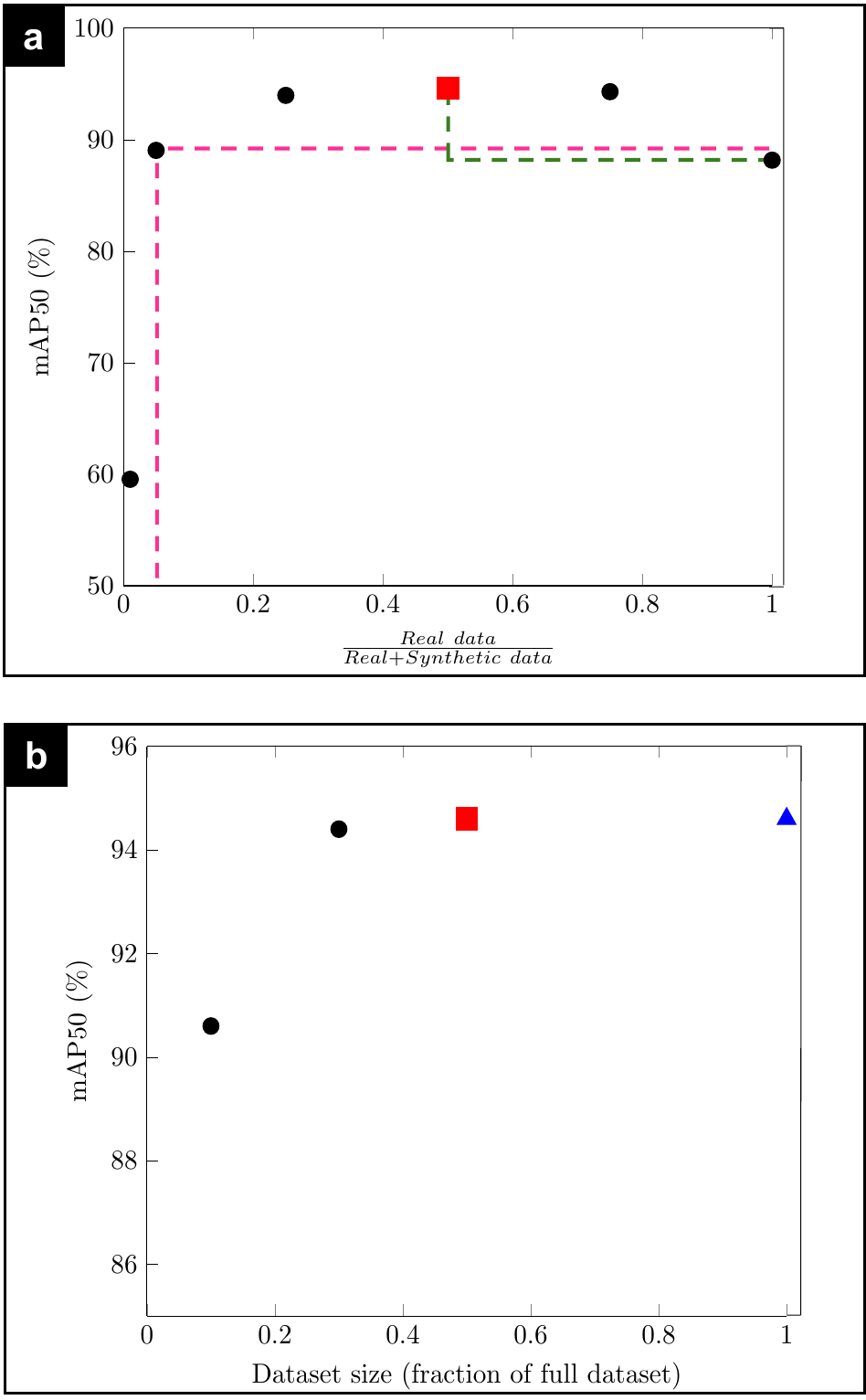}
    \caption{(a) Model performance with respect to different ratios of real and synthetic training data, with constant total dataset size. The pink dashed line highlights a similar level of performance for models trained on datasets that are the same size but comprised of $100\%$ real data and of only $5\%$ real data and $95\%$ synthetic data. The green dashed line highlights performance improvement for $50\%$ real + $50\%$ synthetic data over $100\%$ real data. (b) Model performance with respect to increasing dataset size, keeping a constant real:synthetic ratio of $0.5$. Red square corresponds to the same experiment in (a) and (b) and blue triangle in (b) corresponds to the Real + Syn v0 (Initial) model in Table 1 in the main paper.}
    \label{fig:ratios}
\end{figure}

After adding the initial synthetic data model v0 to the training dataset, we seek to establish the affect on model performance of two main factors: 
\begin{enumerate}
\item{The ratio of real and synthetic data for fixed dataset size.}
\item{The total volume of images (real and synthetic combined) in the training dataset on model performance.}
\end{enumerate}

First, we consider the effect of the ratio of real and synthetic data for fixed dataset size. This tells us the impact of indirect replacement of real images by synthetic images on the model performance, for this dataset. We then use this optimal ratio of real and synthetic data as the training ratio for all of remaining experiments. As we want to use as much of the dataset as possible we use the maximum amount of real images in the dataset to constrain the dataset size. As there are $9000$ real images, our dataset size is also $9000$ images. 
In Supplementary Fig. \ref{fig:ratios}a, we plot the mAP50 scores for models trained on different ratios of real and synthetic data. We observe that mixing just $5\%$ of the real dataset with the synthetic data can provide a similar performance to using all of the real dataset (highlighted by the pink dashed line). Peak performance is achieved at an approximately equal ratio of real and synthetic data (highlighted by the green dashed line). This plot shows that significant performance gains can be achieved through addition of synthetic data to the training set, even without increasing the dataset size. 

Next, assuming that the optimal ratio of the real to synthetic images is constant (at $0.5$) as a function of dataset size, we vary the dataset size to establish the effect of dataset size on model performance. The full dataset size is $18000$ images, $9000$ real images and $9000$ synthetic images. Supplementary Fig. \ref{fig:ratios}(b) shows that increasing the dataset size provides increasing performance up to $50\%$ of the full dataset, and then it saturates for larger dataset sizes.

\section{Angular breakdown of target misclassification and SHAP plots}

At the end of Section 2 in the main paper, we note that the comparison of saliency maps to identify unique and common features between two classes must also account for variation
across each class. In this dataset we have a natural dimension of variation (orientation of the vehicle), and so we do not need to cluster the samples to explore the dimension of variation. To identify which orientations of the target vehicle are being confused with the misclassified vehicle, we first provide a breakdown in segments of $5^{\circ}$ for the target misclassification between the SUV and BTR70 (e.g. the example shown in Fig. 5a-c in the main paper) and the misclassification from the ZSU23 to the BTR70 (e.g. see Fig. 5d-f in the main paper). By providing this angular breakdown we motivate the identification of how confusion between the two different vehicles was identified to be aligned with opposite orientations for the SUV at $\theta\in [70^{\circ}, 75^{\circ}]$ to the BTR70 at $\theta\in [285^{\circ},290^{\circ}]$ and how similar orientations for the ZSU23 at $\theta\in [105^{\circ},110^{\circ}]$ to the BTR70 at $\theta\in [125^{\circ},130^{\circ}]$, as described in the main paper.

\begin{figure*}[ht]
    \centering
    \includegraphics[scale=0.42]{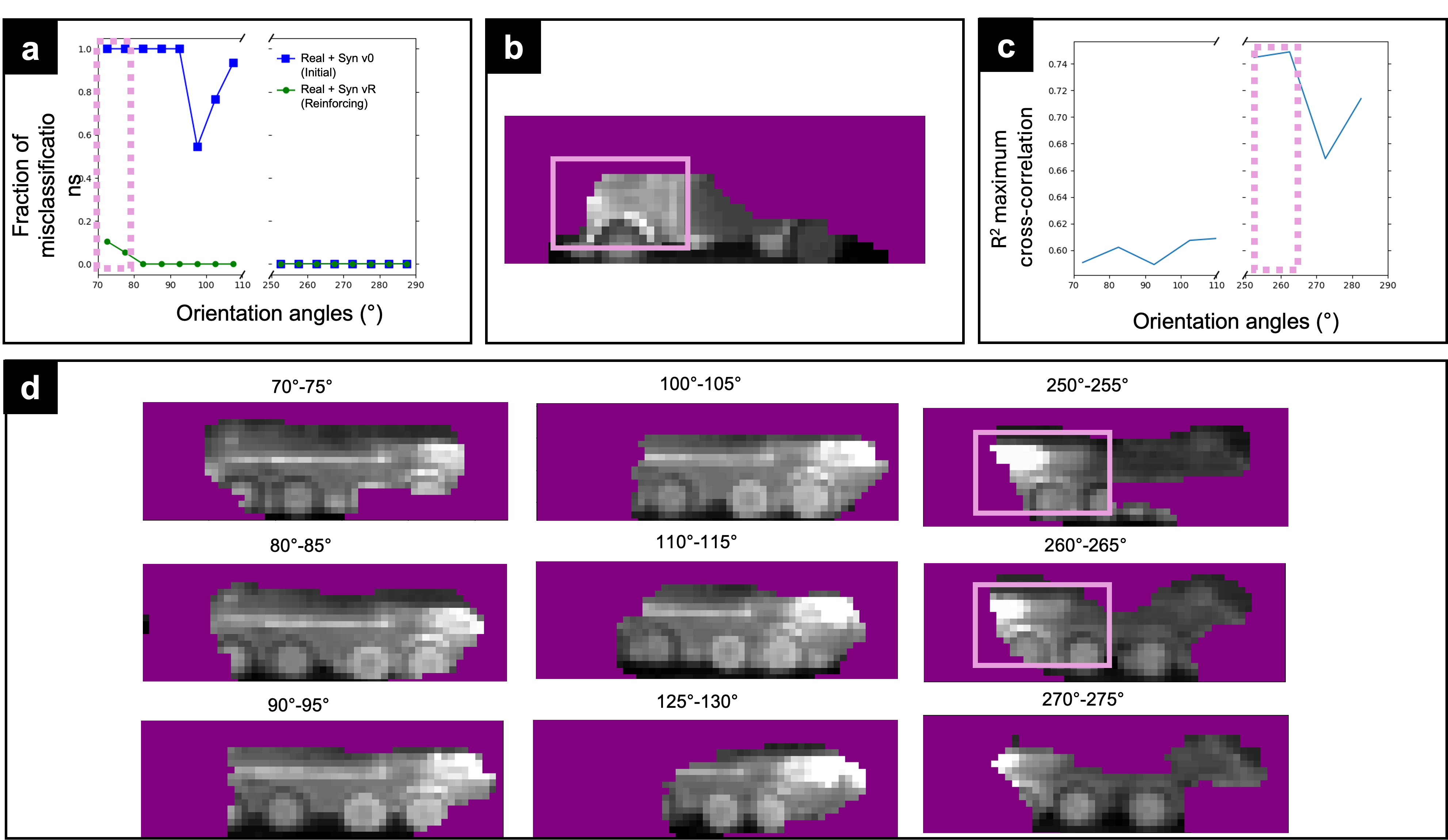}
    \caption{(a) Fraction of the  misclassifications of the SUV as BTR70 for each $5^{\circ}$ over the test dataset. The blue line with squares corresponds to synthetic model v0 before the modification and the green line with circles corresponds to synthetic model vR, after the reinforcing modification. The pink dashed rectangle indicates the orientation range considered for the misclassifications. (b) The average SHAP plot for the SUV misclassifications to BTR70 at $70^{\circ}-75^{\circ}$. (c) Maximum correlations of SUV misclassification SHAP plot in panel (b) with BTR70 correct classifications at different orientations. This helps to determine which subset (of orientations) of BTR70 the SUV is being misclassified as (highlighted by the pink dashed rectangle).(d) Average SHAP plots for the correct classifications of BTR70 at all orientations. The regions of common features associated with misclassification are highlighted by pink rectangles in panels (b) and (c)}
    \label{fig:SUV}
\end{figure*}

Supplementary Fig. \ref{fig:SUV}a shows the breakdown of the misclassifications of the SUV as the BTR70 as a function of orientation. We plot the fraction of misclassifications for this specific misclassification only $(\frac{M_{SUV \rightarrow BTR70}}{C_{SUV}+M_{SUV \rightarrow BTR70}})$ vs. orientation $(\theta)$, where $M_{SUV \rightarrow BTR70}$ is the number of misclassifications of the SUV as the BTR70 and $C_{SUV}$ is the number of correct classifications of the SUV. Supplementary Fig. \ref{fig:SUV}a shows that the fraction of misclassifications for $5^{\circ}$ segments (blue line) is consistently high in the range $70^{\circ}-110^{\circ}$, whilst consistently low in the range $250^{\circ}-290^{\circ}$. This suggests that something on the left side of the vehicle is causing confusion with the BTR70.

In this case, there are multiple segments with an equally high fraction of misclassifications and so we make an arbitrary choice of the segment $70^{\circ}-75^{\circ}$. To identify the likely source of confusion between the SUV in this orientation and the BTR70, we first plot the average SHAP contributions of the misclassification for SUV from the segment $70^{\circ}-75^{\circ}$ in Supplementary Fig. \ref{fig:SUV}b. We then compare this image with the average SHAP contributions from the correct classifications of the BTR70 over an evenly spaced selection of orientations (Supplementary Fig. \ref{fig:SUV}c). The pink boxes in the figure for segments $250^{\circ}-255^{\circ}$, $260^{\circ}-265^{\circ}$, $270^{\circ}-275^{\circ}$ and $280^{\circ}-285^{\circ}$ all show similar features in the vicinity of the wheel arch of the vehicles in a similar location, which is the most likely cause of confusion. These are the patches of common features in Fig. 5b from the main paper. After a separate unique feature on the SUV is identified and reinforced through modifying the synthetic data (see discussion in main paper), the fraction of misclassifications as a function of orientation over the left side of the vehicle is reduced over the entire range as demonstrated by the green line in Supplementary Fig. \ref{fig:SUV}a. 

Similarly for the misclassification between the ZSU23 and the BTR70 from Fig. 5d-f in the main paper, we plot the misclassifications for just the ZSU23 as the BTR70 as a function of orientation in Supplementary Fig. \ref{fig:ZSU23}a. The segments with largest ratio of misclassifications are $105^{\circ}-110^{\circ}$ and $270^{\circ}-275^{\circ}$. As the volume of misclassifications from the left side of the vehicle are slightly higher than the right side we use the $105^{\circ}-110^{\circ}$ segment to identify common features. The average SHAP values for the ZSU23 for this segment are plotted in Supplementary Fig. \ref{fig:ZSU23}b, whilst panel (c) shows the average SHAP values for correct classifications of the BTR70 for all orientations. The orange box over segments $70^{\circ}-75^{\circ}$ to $125^{\circ}-130^{\circ}$ for the BTR70 highlights similar SHAP areas of model focus and image gradients around the rear-left side of the vehicle and wheel arches that are being confused with the rear-left side of the ZSU23 in Supplementary Fig. \ref{fig:ZSU23}b. We identify this as the common feature causing confusion. For this example in the main paper we focus on making disruptive modifications to common features, so we therefore make modifications to our synthetic model of the ZSU23 in this region (as discussed in Section 5.1 in the main paper). The green line with circles in Supplementary Fig. \ref{fig:ZSU23}a demonstrates the performance improvement of the model versus the original blue line with squares, where the misclassifications on the left side of the vehicle have been almost completely removed, whilst on the right side of the vehicle they have also been significantly reduced. Although we did not modify this side of the vehicle this is likely due to subtle changes in high level features recognised over the whole model for this vehicle. 

\begin{figure*}[ht]
    \centering
    \includegraphics[scale=0.42]{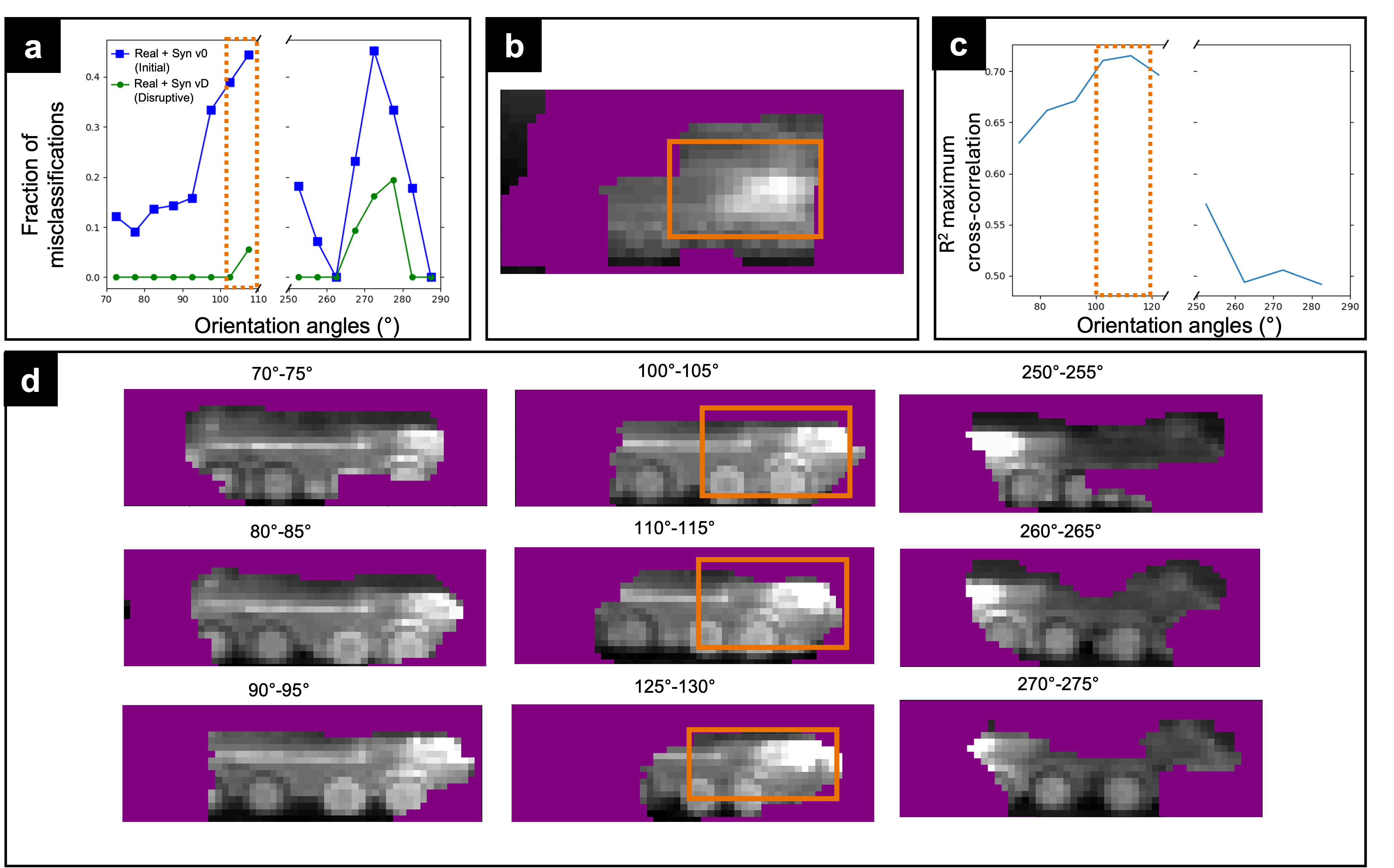}
    \caption{(a) Fraction of the  misclassifications of the ZSU23 as the BTR70 for each $5^{\circ}$ over the test dataset. The blue line with squares corresponds to synthetic model v0 before the modification and the green line with circles corresponds to synthetic model vD, after the disruptive modification. The orange dashed rectangle indicates the orientation range considered for the misclassifications. (b) Average SHAP plot for the ZSU23 at $105^{\circ}-110^{\circ}$. (c) Maximum correlations of ZSU23 misclassification SHAP plot in panel (b) with BTR70 correct classifications at different orientations. This helps to determine which subset (of orientations) of BTR70 the ZSU23 is being misclassified as (highlighted by the orange dashed rectangle). (d) Average SHAP plots for the BTR70 at all orientations. The regions of common features associated with misclassification are highlighted by orange rectangles in panels (b) and (c)}
    \label{fig:ZSU23}
\end{figure*}

\section{Principal Component Analysis of Train-Test Split}
\label{sec:pca}

\begin{figure}
    \centering
    \includegraphics[scale=0.6]{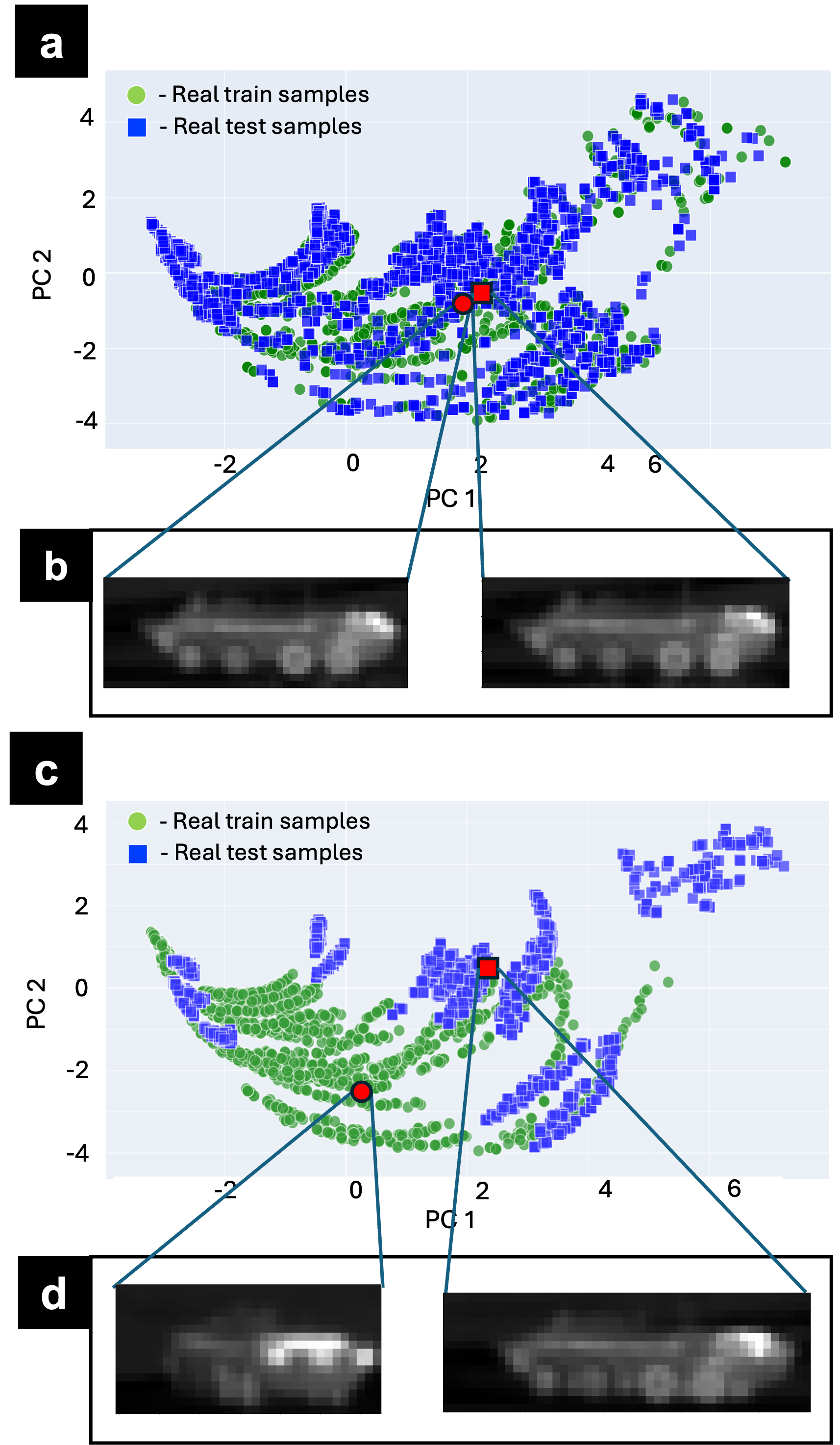}
    \caption{Scatter plot of PCA dimensional reduction to two dimensions of pixels within bounding boxes over the train and test datasets. (a) PCA plot for striped train-test data split. (b) Illustration of two almost identical samples from the train and test datasets, marked by the red circle and red square, respectively. (c) PCA plot for our train-test data split. (d) Illustration of two samples from our train and test dataset split, sampled from the same class (BTR70) to select the closest vehicle orientations between the two datasets, i.e., $160^{\circ}$ from train and $110^{\circ}$ from test. The samples are separated significantly in PCA space, suggesting less likelihood of data leakage between the train and test datasets.}
    \label{fig:pca}
\end{figure}

On inspection of the dataset we noticed that there is considerable similarity between successive video frames. To highlight the potential issue of data leakage between train and test sets we use principal component analysis (PCA) as a first order method to highlight the similarity of train and test samples in principal component space from the two largest principal components, under a train-test split similar to those observed in the literature of this dataset \cite{Priddy2023ExplorationsIT, 2019SPIE10988E..08M, Sami2023DeepTT, 2024SPIE13083E..17O, 9438143}. Typical train-test splits use striped splitting, where every 1-in-10 frames is used for the train subset, or random sampling.   
Supplementary Fig \ref{fig:pca}a shows the train data (green) and test data (blue) or the striped train-test split. There is a large overlap over the entirety of the principal component space between the train and test datasets for this striped splitting. Furthermore, Supplementary Fig. \ref{fig:pca}b shows examples of train (left) and test (right) vehicles given by the red circle and red square in (a) respectively. The vehicles in the striped split in panel (b) are visually almost identical and are located close to each other in both real space and in PCA space. This example demonstrates the likelihood that the train-test split contains large amount of domain leakage.

We use this as motivation for constructing a train-test split with significantly less domain leakage identified by the distinct clusters in PCA space. Our train-test split is represented in Fig. 3 in the main paper, and shown here in Supplementary Fig. \ref{fig:pca}c, where the train and test datasets remain distinct. The example vehicles in our train-test split in panel (d) are orientated at $160^{\circ}$ (train) and $110^{\circ}$ (test), which is the closest two orientations of this vehicle between our train and test datasets (e.g. see Fig. 3 in main paper, left and middle panels). These examples both differ significantly in real space and in PCA space. This indicates that our train-test split is unlikely to contain domain leakage. 

